\documentclass{article}
\usepackage{mathpazo}
\usepackage[margin=1in]{geometry}
\usepackage[style=alphabetic,natbib=true]{biblatex}
\usepackage{microtype}
\usepackage{graphicx}
\usepackage{subfigure}
\usepackage{booktabs} 
\usepackage{hyperref}

\usepackage{amsmath}
\usepackage{amssymb}
\usepackage{mathtools}
\usepackage{amsthm}

\usepackage{bbold}
\usepackage{bm}

\DeclareMathOperator*{\argmax}{arg\,max}
\newcommand*{\escape}[1]{\texttt{\textbackslash#1}}
\usepackage{booktabs}
\usepackage{multirow}
\usepackage{algorithm}
\usepackage{algpseudocode}
\usepackage[dvipsnames]{xcolor}
\usepackage{longtable}

\usepackage[capitalize,noabbrev]{cleveref}

\theoremstyle{plain}

\theoremstyle{definition}

\theoremstyle{remark}

\usepackage[textsize=tiny]{todonotes}

\title{In-context Example Selection with Influences}
\author{
    Tai Nguyen \\
    {\normalsize \textsc{taing@seas.upenn.edu}}\\
    University of Pennsylvania
    \and
    Eric Wong \\
    {\normalsize \textsc{exwong@cis.upenn.edu }}\\
    University of Pennsylvania
}
\date{}

\begin{document}

\maketitle

\begin{abstract}
In-context learning (ICL) is a powerful paradigm emerged from large language models (LLMs). Despite its promises, ICL performance is known to be highly sensitive to input examples. In this work, we use \textit{in-context influences} to analyze few-shot ICL performance directly from the in-context examples. Our proposed influence-based example selection method can identify both positive and negative examples, outperforming several baselines when evaluated on 9 SuperGLUE tasks. Our analysis uncovers up to a $16.3\%$ performance gap between using the most negative in-context examples compared to the most positive. In a case study, we apply our influence-based framework to quantify the phenomena of recency bias in example ordering for few-shot ICL.\footnote{Our code is released at \url{https://github.com/DebugML/incontext_influences}.}
\end{abstract}
\section{Introduction}
\label{introduction}

Large language models (LLMs) such as GPT-3 have recently become capable of \textit{in-context learning} (ICL) \cite{brown_language_2020}. In ICL, users provide the model with a few labeled examples as input before asking the model to make a prediction on a new example. This paradigm has enabled the rapid adaptation of LLMs to new tasks without requiring any modifications to the model.

ICL has several advantages over traditional learning paradigms. First, the ability to do few-shot learning directly reduces the need for human-labeled data. Second, in contrast to other popular training paradigms such as finetuning a pretrained model \cite{Radford2019LanguageMA,devlin-etal-2019-bert}, ICL enables inference without any gradient updates. Lastly, ICL also displays amazing versatility through different modes of prompting. Recent work shows that GPT-3 can do step-by-step reasoning when being demonstrated a few examples containing reasoning \cite{wei2022chain, nye2022show, lyu2023faithful}.

Despite these promises, ICL performance is known to be highly variable. In particular, ICL volatility has been linked to 
biases such as the order of the examples \cite{lu-etal-2022-fantastically}, their templates \cite{lu-etal-2022-fantastically,kumar_reordering_2021}, and example selection \cite{liu-etal-2022-makes}. Various mitigation methods were proposed to address this brittleness, such as  model calibration \cite{pmlr-v139-zhao21c} and template engineering \cite{liu-prompting-2022}.

Given that not all in-context examples are equal, several others have focused on finding more optimal prompts. \citet{liu-etal-2022-makes} proposes a distance-based selection method, using semantic similarity to the validation query to rank candidate examples. \citet{gonen_perplexity_2022} finds a strong negative correlation between example perplexity and task performance. Similarly, \citet{chen_sensitity_2022} suggests a sensitivity-based selection method which perturbs examples and chooses the ones with more robust predictions. While these methods have varying effectiveness, there lacks a consensus on which of these signals are most important in ICL.

Motivated by this problem, our paper studies the relationship between influences and ICL, to better understand and quantify the impact of examples on ICL. Influences naturally lend to an offline example selection method that directly measures the effect of examples on ICL performance. 
In particular, we use \emph{in-context influences} to measure and rank the impact of in-context examples on task performance. The framework can be customized to study different aspects of ICL, such as optimizing for the best classification accuracy or quantifying the impact of position. 

On 8 language models and 9 natural language tasks, we demonstrate the efficacy of \emph{influence-based example selection} in ICL at estimating the effect of training examples on downstream performance.
We find that in-context influences outperform all other selection baselines at estimating ICL performance in both positive and negative selections. In-depth analysis exposes a significant gap between the most positive and most negative examples. For example, constructing prompts from the top influence bin improves ICL performance by up to $16.3\%$ over the bottom influence bin on LLaMA-13B.

Overall, our contributions are as follows:
\begin{itemize}
    \item We study in-context influence as a metric for selecting and analyzing in-context examples in few-shot ICL. In both positive and negative selections, our method outperforms several baselines at estimating ICL performance.
    \item We demonstrate a substantial performance gap between positively and negatively influential examples. Our framework quantifies this gap, and further confirms the variability of example-selection in ICL. 
    \item While we focus on classification accuracy, our framework generalizes to any combination of performance metric, model, and task. For example, we leverage our framework to quantify emergent phenomena in LLMs, such as the impact of recency bias in example ordering.
\end{itemize}

\section{In-context Influences}
\label{section:definition}
\begin{figure*}[t]
\centering
\centerline{\includegraphics[width=1.0\textwidth]{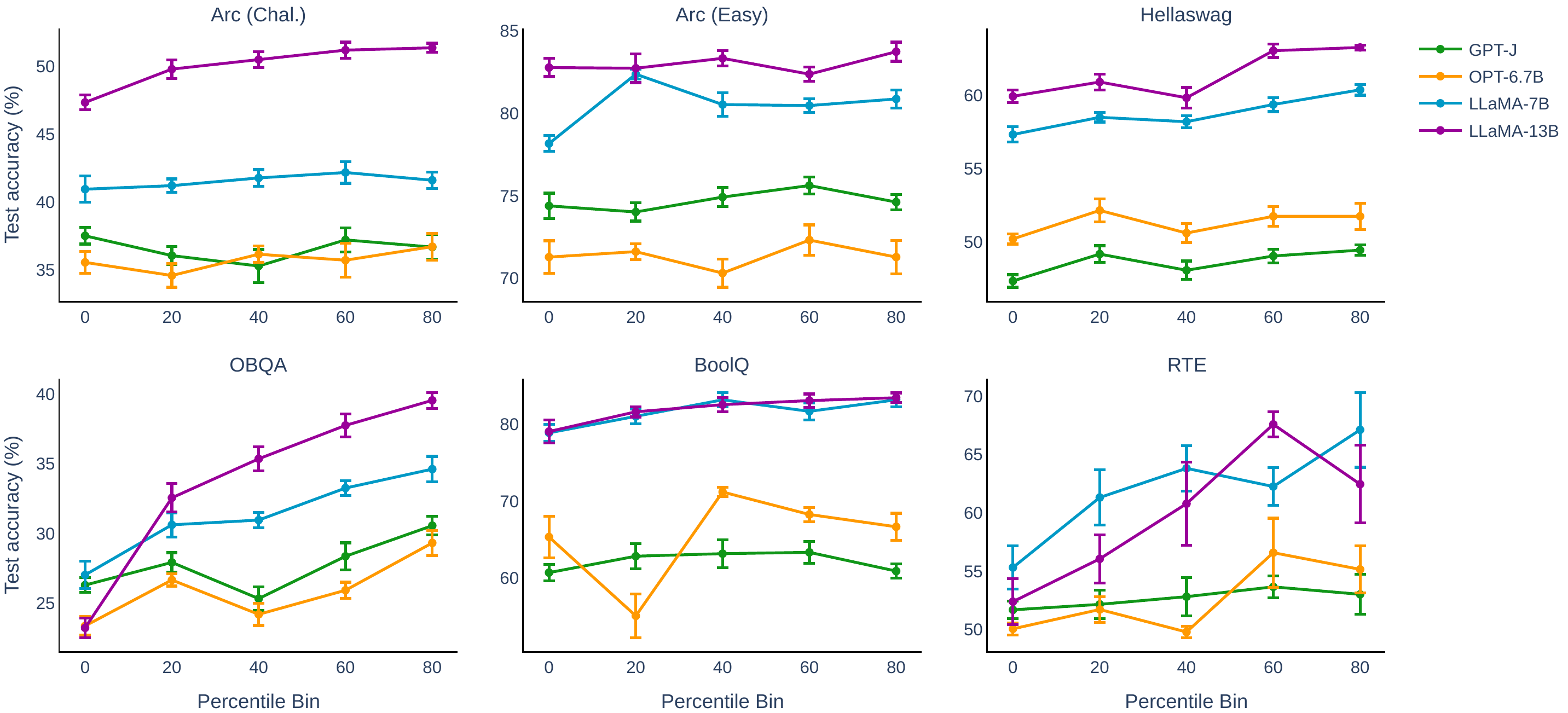}}
\caption{Test accuracy increases when examples are selected in increasing in-context influence percentile bins.}
\label{fig_influence_bins}
\end{figure*}

A variety of methods have been developed to understand how training data affects model performance. To estimate this effect, some methods use gradient information \cite{koh2017understanding, koh_groups_NEURIPS2019,han_explaining_2020, tracin} while others retrain models on subsets of the training data \citep{pmlr-ghorbani,ilyas_datamodels_2022}. 
These methods all aim to quantify how a training example affects the prediction of a test example after training. Inspired by these frameworks, our goal is to trace how ICL performance depends on the in-context examples and calculate the corresponding influences.

Our setup follows the retraining-based influence frameworks, which have two main steps. Let $S$ be a training set, and let $f(S)$ be the validation performance after training on a dataset $S$. Retraining-based influences first collect a ``dataset'' of $M$ training runs $\mathcal D = \{(S_i, f(S_i)\}_{i=1}^M$ where $S_i\subseteq S$ are random subsets of the original training dataset. The second step is to use this dataset to estimate the influence of each example $x\in S$, e.g. by learning a linear mapping \cite{ilyas_datamodels_2022}.

\paragraph{Influences in $k$-shot prompting.} To compute influences for in-context examples, we leverage the following key observation: in ICL, ``training'' a model on a subset $S'$ reduces to prompting the model on a sequence containing $S'$. Consequently, constructing the dataset $\mathcal D$ of training runs for ICL requires no gradient updates and is as costly as computing forward passes through the model. This drastically reduces the cost of calculating retraining-based influences, and can be calculated with only query-access to the model. 

Specifically, for the first step, we construct the dataset of training runs $\mathcal D$ by performing $k$-shot prompting with subsets $S'\subseteq S$ where $|S'|=k$. For a fixed subset $S'$, the performance of the resulting prompt containing $S'$ is measured with a validation \textit{query} appended to the end of the prompt. We repeat this inference over the entire validation set to compute the metric $f(S')$, which measures the validation performance after prompting with $S'$. This metric can be any evaluation method suitable for a natural language task---here, we focus on classification accuracy. We repeat this process over multiple random subsets $S'\subseteq S$ until each example in $S$ has been seen in multiple prompts, resulting in a dataset of prompting runs $\mathcal D = \{(S_i, f(S_i)\}_{i=1}^M$.

In the second step, we calculate the influence of each in-context example. 
We define the in-context influence $\mathcal{I}(x_j)$ as the effect of an example $x_j$ on few-shot ICL performance. In other words, the influence is the difference between the average performance of prompts including $x_j$ and the average performance of prompts omitting $x_j$. More formally, this can be written as:

\begin{equation}
    \label{eq:influence}
    {\mathcal{I}(x_j)=\frac{1}{N_j}\sum_{S_i:x_j\in S_i} f(S_i)} - {\frac{1}{M-N_j}\sum_{S_i:x_j\notin S_i} f(S_i)}
\end{equation}
where $S_i$ is a specific uniformly sampled subset, $M$ is the number of total subsets used to estimate influences, $N_j$ is the total number of subsets containing example $x_j$, and $f(S_i)$ is the performance metric when evaluated on the validation set. When $f$ measures validation performance, a higher score for $\mathcal{I}(x_j)$ corresponds to a higher average improvement in validation performance when including $x_j$ in the prompt, analogous to the meaning of influences in the classic, non-prompted setting.

As the number of collected subsets grows, estimates of in-context influences become more accurate. A sufficiently large $M$ is one with good \textit{coverage} for each example---this means that each $x_j \in S$ is seen multiple times. In our experiments, each $x_j$ gets seen at least 30 times on average.

\begin{algorithm}[t]
   \caption{Influence-based Example Selection}
   \label{alg_influences}
{\textbf{Input:} Language model $\texttt{LLM}$, training set $S=\{X_j=(x_j; y_j)\}^N_{j=1}$, validation set $V$, test set $T$, performance metric $f$, number of in-context examples $k$ (hyperparameter), and $M$ number of total subsets (hyperparameter).}

\textbf{Step 1:} Subset collection
\begin{algorithmic}[1]
   \For {\textbf{$i=1$ to $M$}}
       \State Randomly select subset $S_i\subseteq S$, where $|S_i|=k$
       \State Compute $f(S_i)$ over $V$
       \State Store the pair $\{S_i, f(S_i)\}$
    \EndFor
\end{algorithmic}

\textbf{Step 2:} Calculate example influence
\begin{algorithmic}[1]
   \For {$X_j \in S$}
       \State Compute $\mathcal{I}(X_j)$ following Equation~\ref{eq:influence}
    \EndFor
\end{algorithmic}

\textbf{Step 3:} Inference
\begin{algorithmic}[1]
       \State Select $k$ examples $\{X_1'\dots X_k'\}\subset S$ with the largest influence scores $\mathcal{I}(X_{j}')$
       \State Construct $C = [X_{1}', ..., X_{k}']$ in any ordering
       \State $\hat{y}_{test} = \texttt{LLM}(C; x_{test})$
\end{algorithmic}
\end{algorithm}

\paragraph{Influence-based Example Selection.} We use the proposed in-context influences to identify highly impactful in-context examples. 
Specifically, we can use the top influential examples to create the ``best'' prompt for ICL (with respect to the influence scores). On the converse, we can also use the bottom influential examples to create the ``worst'' performing prompt for ICL. In summary, to do example selection for ICL, we carry out the following steps:
\begin{enumerate}
    \item Prompt the model on random training subsets and measure validation performance to create the dataset of prompting runs $\mathcal D$.
    \item Calculate the in-context influence $\mathcal I(x_j)$ for each example $x_j \in S$ following Equation \ref{eq:influence} using $\mathcal D$.
    \item Select $k$ examples with the most positive influences to use for $k$-shot prompting. The examples can be arranged in any ordering.
\end{enumerate}
A summary of the entire pipeline is shown in Algorithm~\ref{alg_influences}.

\subsection{Cost Analysis \& Hyperparameters}

\paragraph{Training cost.} Retraining-based influence frameworks \citep{ilyas_datamodels_2022,pmlr-ghorbani} can require training hundreds of thousands of models. This is necessary to collect a sufficiently large enough dataset $\mathcal D$ to accurately estimate influences. In contrast, the cost of computing in-context influences is relatively cheap, as we do not need to train an end-to-end model. Instead of training, we simply prompt the LLM using a randomly sampled $S'$ from original training set $S$. Thus, the complexity of calculating the validation performance from a sampled subset is proportional to a forward pass through the LLM. 
\paragraph{Size of subsets.} Our method has one parameter $k$, which controls the size of the random subsets $S'\subseteq S$ from which $\mathcal D$ is constructed. For ICL, $k=|S'|$ corresponds to the number of in-context examples given in the prompt. Unlike in the traditional setting, the context window length limit enforces a hard upper limit on the number of examples an LLM can be trained on via prompting. These context windows are typically limited to 2048 characters.  

Taking the context window into account, we select $k$ to be the maximal number of examples that can be inserted into the context window. The value of $k$ can vary by different choices of model (context window size)
and the lengths of the individual examples in a dataset. Since the number of shots can impact ICL performance, we keep a consistent $k$ for each model and task. Table~\ref{tab_datasets} provides the precise $k$ value associated with each task.
\section{Experiments}
\label{section:experiments}
We conduct experiments to obtain in-context influences for 72 combinations of natural language tasks and LLMs. The goal is to a select a set of good ICL examples by running influences on the Dev set. At test time, such a set requires \textit{no further} modification or computation.

\paragraph{Datasets.} We choose 9 datasets for our study, 5 of which are binary classification tasks and 4 are multi-choice tasks.\footnote{Table~\ref{tab_datasets} in the Appendix summarizes our choices and subsamples.} These datasets cover a wide range of common natural language tasks, including textual entailment (RTE), question-answering (PIQA), and text summarization (BoolQ). Each example instance has a definitively correct answer, making them convenient to be evaluated through classification accuracy. Beyond acquiring the original data, we subsample Train/Dev/Test sets with 400/200/500 in-context examples for each task.

\paragraph{Models.} Our work uses 3 publicly available LM families, including \textbf{LLaMA} (7B, 13B) \cite{touvron2023llama}, \textbf{OPT} (6.7B, 13B, 30B) \cite{zhang_opt_2022}, and \textbf{GPT-Neo} (GPT-J 6B, GPT-NeoX 20B) \cite{black_gpt-neox-20b_2022}.

\paragraph{Prompt format.} For each task, we follow the same template used in \citet{brown_language_2020}. If such a template is not accessible, we construct our own templates by keeping them minimal without intensive prompt engineering \cite{perez2021true}. Table~\ref{tab_prompts} in the Appendix provides full prompt details.

\paragraph{K-shot selection.} We select examples uniformly by their label class. If a multi-choice task has 3 options, each option would make up roughly one-third of the examples. This balance prevents majority label bias \cite{pmlr-v139-zhao21c} from skewing the model's inference.

\paragraph{Inference details.} There are multiple ways to do inference on multi-choice tasks with autoregressive models \cite{surface}. We follow one popular approach, which ranks all possible continuations to a prompt and chooses the continuation with the highest log-likelihood. Thus, given a prompt $x_{0:m}$ and a possible prompt continuation $x_{m:n}$, the score for $x_{m:n}$ can be defined as:

$$\ell(x_{m:n}) = \sum_{j=m}^{n} \log \mathbb{P}(x_j | x_{0:j})$$

where $\mathbb{P}(x_j|x_{0:m})$ is the likelihood of token $x_j$ given the preceding context tokens $x_{0:j}$. The prediction is then defined as the most likely continuation, $\argmax_{x_{m:n}} \ell(x_{m:n})$. We do not perform any token length or answer normalization tricks \cite{brown_language_2020}.

\subsection{Influence-based methods}
\label{section:infl-baselines}
In our main results, we evaluate the effectiveness of influence-based example selection using the following strategies:
\begin{enumerate}
    \item \textbf{Influence ($+$/$-$).} We select examples with the most positive or negative influence scores according to Algorithm~\ref{alg_influences}. If influence estimates are meaningful, we would expect examples with positive influences to perform the best among all baselines, while examples with negative influences would have the poorest performance.

    \item \textbf{In-context datamodels.} We consider an alternative influence-based example selection based on the datamodels \citep{ilyas_datamodels_2022} framework that we adapt for ICL. Specifically, we fit a linear model\footnote{Following \citet{ilyas_datamodels_2022}, we choose an L1 regularized regression with penalty $\lambda=0.0001$.} $g_\theta$ on the dataset $\mathcal D$ of input-output pairs from Section~\ref{section:definition} to predict validation performance:
    $$g_\theta(S') = \theta \cdot \mathbf{1}_{S'}^T  + \theta_0$$
    
    where $S'\subseteq S$ is an example subset and $\mathbf{1}_{S'}$ is an indicator vector with the dimension of the training set $S$. A value of 1 at position $i$ indicates that the example $i$ is included in the subset $S'$ and a value of 0 means otherwise. Following the datamodels framework, we can treat the parameters $\theta$ as influence estimates, and select in-context examples based on these estimates. Note that $\theta$ has a close connection to in-context influences as they both use the same training set $\mathcal D$, but in-context datamodels assumes a linear model.
\end{enumerate}

\subsection{Non influence-based methods}
\label{section:non-infl-baselines}
We compare influence-based example selection methods described in the previous section to the following baselines, which optimize various metrics for selection.

\begin{enumerate}
    \item \textbf{Random.} We randomly select a set of in-context examples for inference.
    \item \textbf{Best set.} We select the best set of examples observed during the collection of training runs $\mathcal{D}$ for computing in-context influences.
    \item \textbf{One-shot.} We do one-shot prompting ($k=1$) on each Train example and rank them by their accuracy on the Dev set. One-shot selection assumes that we can extrapolate the performance of one-shot prompting to the few-shot setting.
    \item \textbf{Semantic similarity.} Examples close to the test queries in the embedding space can substantially improve ICL performance on semantic parsing tasks \cite{liu-etal-2022-makes}. We search for a set of examples with the closest distance to Dev set, then applying them on the unseen Test set. We use RoBERTa-large \cite{roberta} sentence encoder implemented by \citet{sbert}.
    \item \textbf{Perplexity.} Perplexity measures the degree of uncertainty of the LLM when generating new tokens, where a lower perplexity means a higher confidence on the example. On perplexity, \citet{gonen_perplexity_2022} has linked prompt confidence to good ICL performance. We follow this insight to select examples based on their individual perplexity, which is more computationally friendly than calculating full prompt perplexity across many different example combinations. 
\end{enumerate}

After selecting a set of $k$ examples using the proposed baselines, we construct the prompt by ordering examples randomly. We compute Test accuracy and rank all baselines by single-task accuracy. We aggregate the ranks of each method my taking their average for both positive and negative example selection. In the main results, averages and standard errors are reported over 7 seeds.

\begin{table*}[t]
\caption{Positive example selection methods on OPT-30B and their overall Rank Aggregation.}
\label{tab_mainresults-pos}
\vskip 0.1in
\centering
\resizebox{\textwidth}{!}{
\begin{tabular}{llllllllllr}
\toprule
& \multicolumn{9}{c}{OPT-30B} & All Models\\
\cmidrule(lr){2-10} \cmidrule(lr){11-11}
& \multicolumn{5}{c}{Binary Classification (Acc. $\uparrow$)} & \multicolumn{4}{c}{Multi-choice (Acc. $\uparrow$)} & Rank Agg ($\downarrow$) \\
\cmidrule(lr){2-6}\cmidrule(lr){7-10} \cmidrule(lr){11-11}
~ &                               PIQA &                             BoolQ &                                RTE &                                WIC &                               WSC & ARC-c & ARC-e & HS &                              OBQA &   All Tasks \\
\midrule
Perplexity (+)    &           76.8\scalebox{0.5}{0.0} &           72.7\scalebox{0.5}{0.2} &           61.9\scalebox{0.5}{0.3} &           53.5\scalebox{0.5}{0.2} &           43.5\scalebox{0.5}{0.6} &               40.3\scalebox{0.5}{0.1} &              76.3\scalebox{0.5}{0.1} &              56.6\scalebox{0.5}{0.0} &           28.5\scalebox{0.5}{0.1} &  4.59 \\
Random            &           77.0\scalebox{0.5}{0.0} &           71.1\scalebox{0.5}{0.2} &           63.2\scalebox{0.5}{0.2} &           54.8\scalebox{0.5}{0.1} &           49.1\scalebox{0.5}{0.5} &               41.5\scalebox{0.5}{0.1} &              76.0\scalebox{0.5}{0.1} &              55.4\scalebox{0.5}{0.1} &           29.6\scalebox{0.5}{0.1} &  4.37 \\
Similarity (+)    &           77.7\scalebox{0.5}{0.1} &           70.1\scalebox{0.5}{0.4} &           63.9\scalebox{0.5}{0.1} &           53.3\scalebox{0.5}{0.1} &           57.1\scalebox{0.5}{0.7} &      42.0\scalebox{0.5}{0.1} &              76.2\scalebox{0.5}{0.1} &              56.7\scalebox{0.5}{0.0} &           29.3\scalebox{0.5}{0.0} &  4.33 \\
One-shot (+)      &           77.5\scalebox{0.5}{0.0} &           76.5\scalebox{0.5}{0.1} &           52.4\scalebox{0.5}{0.1} &           51.1\scalebox{0.5}{0.2} &  \textbf{61.6\scalebox{0.5}{0.0}} &               41.5\scalebox{0.5}{0.0} &              76.1\scalebox{0.5}{0.1} &              56.6\scalebox{0.5}{0.1} &           31.2\scalebox{0.5}{0.0} &  4.24 \\
Best set          &           76.9\scalebox{0.5}{0.0} &           72.6\scalebox{0.5}{0.0} &           64.1\scalebox{0.5}{0.3} &  \textbf{55.1\scalebox{0.5}{0.2}} &           54.8\scalebox{0.5}{0.4} &               40.8\scalebox{0.5}{0.0} &              75.8\scalebox{0.5}{0.1} &              56.1\scalebox{0.5}{0.0} &           31.5\scalebox{0.5}{0.0} &  3.62 \\
\midrule
IC Datamodels (+) &  \textbf{78.1\scalebox{0.5}{0.0}} &  \textbf{77.0\scalebox{0.5}{0.0}} &  \textbf{65.9\scalebox{0.5}{0.1}} &           51.4\scalebox{0.5}{0.2} &           56.4\scalebox{0.5}{0.1} &      \textbf{42.1\scalebox{0.5}{0.0}} &              76.6\scalebox{0.5}{0.0} &     \textbf{58.2\scalebox{0.5}{0.0}} &           31.7\scalebox{0.5}{0.1} &  2.98 \\
Influence (+)     &           78.0\scalebox{0.5}{0.0} &           74.1\scalebox{0.5}{0.1} &           64.6\scalebox{0.5}{0.1} &           52.5\scalebox{0.5}{0.1} &           51.4\scalebox{0.5}{0.3} &               41.6\scalebox{0.5}{0.1} &     \textbf{77.0\scalebox{0.5}{0.0}} &              57.4\scalebox{0.5}{0.0} &  \textbf{33.3\scalebox{0.5}{0.0}} &  \textbf{2.96} \\
\bottomrule
\end{tabular}
}
\end{table*}

\begin{table*}[t]
\caption{Negative example selection methods on LLaMA-13B and their overall Rank Aggregation.}
\label{tab_mainresults-neg}
\vskip 0.1in
\centering
\resizebox{\textwidth}{!}{
\begin{tabular}{llllllllllr}
\toprule
& \multicolumn{9}{c}{LLaMA-13B} & All Models\\
\cmidrule(lr){2-10} \cmidrule(lr){11-11}
& \multicolumn{5}{c}{Binary Classification (Acc. $\downarrow$)} & \multicolumn{4}{c}{Multi-choice (Acc. $\downarrow$)} & Rank Agg ($\downarrow$)\\
\cmidrule(lr){2-6}\cmidrule(lr){7-10} \cmidrule(lr){11-11}
~ &                               PIQA &                             BoolQ &                                RTE &                                WIC &                               WSC & ARC-c & ARC-e & HS &                              OBQA &   All Tasks \\
\midrule
Similarity (-)    &           79.2\scalebox{0.5}{0.0} &           83.2\scalebox{0.5}{0.0} &           58.7\scalebox{0.5}{0.1} &           54.6\scalebox{0.5}{0.2} &           43.9\scalebox{0.5}{0.5} &           51.1\scalebox{0.5}{0.0} &           82.3\scalebox{0.5}{0.0} &           62.1\scalebox{0.5}{0.0} &           37.1\scalebox{0.5}{0.0} &  5.19 \\
Random            &           78.5\scalebox{0.5}{0.1} &           82.6\scalebox{0.5}{0.1} &           61.1\scalebox{0.5}{0.3} &           51.8\scalebox{0.5}{0.2} &           42.9\scalebox{0.5}{0.4} &           50.4\scalebox{0.5}{0.1} &           82.7\scalebox{0.5}{0.0} &           62.5\scalebox{0.5}{0.1} &           35.7\scalebox{0.5}{0.1} &  4.94 \\
Worst set         &           78.8\scalebox{0.5}{0.0} &           79.2\scalebox{0.5}{0.1} &           54.1\scalebox{0.5}{0.2} &           53.3\scalebox{0.5}{0.1} &           45.7\scalebox{0.5}{0.6} &           50.3\scalebox{0.5}{0.1} &           83.0\scalebox{0.5}{0.0} &           62.1\scalebox{0.5}{0.1} &           33.6\scalebox{0.5}{0.1} &  4.37 \\
Perplexity (-)    &  \textbf{74.9\scalebox{0.5}{0.0}} &           82.4\scalebox{0.5}{0.1} &           57.9\scalebox{0.5}{0.1} &           55.4\scalebox{0.5}{0.2} &           42.8\scalebox{0.5}{0.4} &           49.4\scalebox{0.5}{0.0} &  \textbf{81.4\scalebox{0.5}{0.0}} &  \textbf{58.7\scalebox{0.5}{0.0}} &           33.1\scalebox{0.5}{0.1} &  3.69 \\
One-shot (-)      &           78.7\scalebox{0.5}{0.0} &  \textbf{68.2\scalebox{0.5}{0.2}} &           53.9\scalebox{0.5}{0.1} &           53.1\scalebox{0.5}{0.1} &           55.4\scalebox{0.5}{0.7} &           50.0\scalebox{0.5}{0.1} &  \textbf{81.4\scalebox{0.5}{0.0}} &           61.0\scalebox{0.5}{0.0} &           26.1\scalebox{0.5}{0.1} &  2.96 \\
\midrule
IC Datamodels (-) &           78.5\scalebox{0.5}{0.0} &           69.3\scalebox{0.5}{0.3} &  \textbf{50.0\scalebox{0.5}{0.0}} &           51.6\scalebox{0.5}{0.2} &  \textbf{38.9\scalebox{0.5}{0.1}} &           50.0\scalebox{0.5}{0.1} &           82.8\scalebox{0.5}{0.1} &           61.8\scalebox{0.5}{0.0} &  \textbf{22.0\scalebox{0.5}{0.1}} &  3.03 \\
Influence (-)     &           78.6\scalebox{0.5}{0.0} &           68.3\scalebox{0.5}{0.3} &  \textbf{50.0\scalebox{0.5}{0.0}} &  \textbf{50.6\scalebox{0.5}{0.2}} &           39.8\scalebox{0.5}{0.3} &  \textbf{49.3\scalebox{0.5}{0.1}} &           82.4\scalebox{0.5}{0.1} &           61.6\scalebox{0.5}{0.0} &           22.9\scalebox{0.5}{0.1} &  \textbf{2.90} \\
\bottomrule
\end{tabular}
}
\begin{flushleft}
\vskip -0.1in
\rule{0in}{1.3em}$^\dag$\scriptsize Full results for positive and negative example selection methods are provided in Table~\ref{tab_mainresults-pos-full} and Table~\ref{tab_mainresults-neg-full} in the Appendix.
\end{flushleft}
\end{table*}

\subsection{Results}
\label{section:results}
\paragraph{Positive selection.} Table~\ref{tab_mainresults-pos} shows results for all positive example selection baselines.
Overall, influence-based selection methods outperform all non-influence counterparts. Specifically, across all models and tasks, Influence (+) and IC Datamodels (+) frequently select the best set of examples for an average rank of 2.98 and 2.96 (where both methods are considered in the ranking). Best set selection is our next most competitive baseline, with the rest of the methods trailing far behind. For Best set, strong performance on the task WIC (word sense disambiguation) contributes mostly to this success. Likewise, One-shot example selection sees exceptional performance on WSC, but does not work well for other tasks.
We also note that random selection ranks better than Perplexity (+), although the latter outperforms random selection on many tasks (see Table~\ref{tab_mainresults-pos-full} in Appendix).

\paragraph{Negative selection.} Similarly, our influence-based example selection methods can consistently identify low-performing examples. As shown in Table~\ref{tab_mainresults-neg}, results indicate that Influence (-) achieves the highest rank (2.90) among all other methods. Notably, in this context, One-shot slightly outperforms IC Datamodels (-), suggesting that selecting examples based on their individual validation performance can be effective. Compared to positive selection, the ability to pinpoint negative examples is equally meaningful: we can avoid these examples to achieve better ICL performance, or further study them to identify the factors that make them ineffective.

\paragraph{Binary vs. Multi-choice.} Our experiments find influence-based example selection methods to work better on Multi-choice tasks compared to Binary classification tasks. In most models (see Table~\ref{tab_mainresults-pos-full} in Appendix), Influence (+) and IC Datamodel (+) often outperform all other methods on multi-choice tasks, but less so on binary classification tasks. Specifically, for PIQA and WIC, the small gaps between the performance of Influence (-) and Influence (+) suggest that our influence-based method might not have captured example helpfulness well for these tasks. Many factors related to both the model and task could explain such disparity. For one, \citet{pmlr-v139-zhao21c} demonstrates that LLMs can have a strong bias towards selecting certain labels for ICL, which could hinder both model performance and influence attribution. We suspect that these biases can exacerbate when the LLM is asked to choose between binary labels (T/F) compared to many labels in the multi-choice setup. Furthermore, the inconsistent scaling of 3 OPT models on the SuperGLUE benchmark for few-shot ICL could play a factor \cite{zhang_opt_2022}. The capabilities of the models themselves factor majorly into the accuracy improvements of our selection methods.
\section{Analysis}

In this section, we analyze in-context influences across a number of distinct axes. Specifically, we study 
(1) the cost of influence-based example selection,
(2) distinguishing factors between examples with positive and negative influences, (3) influence agreement between model families, and (4) the scaling behavior of influence-based selection across the number of shots. We conclude our analysis with a case study quantifying the effect of recency bias in example ordering. 

\subsection{Cost Comparison}
 Figure~\ref{fig_token_match} compares the cost of influence-based example selection against Best set and One-shot example selections on different tokens budgets for multi-choice tasks.\footnote{We use the GPT-2 Byte Pair Encoding (BPE) tokenizer, which OpenAI also uses for their API pricing.} Recall that computing our in-context influences on more training runs often leads to more accurate influence estimations. Our visualization demonstrates that in-context influences can realize favorable gains over other selection methods at a fraction of the full budget (20M tokens). In fact, in-context influences also scale well beyond this number, while the same effect does not guarantee for Best set. Note that One-shot selection scales linearly by the size of the Train set $S$ and Dev set, while Influence (+) has more flexible scaling depends on compute budget.

\subsection{Do models agree on high-influence examples?}
\label{section:family-agreement}
\begin{figure}[t]
\begin{minipage}[t]{0.47\columnwidth}
    \includegraphics[width=\linewidth]{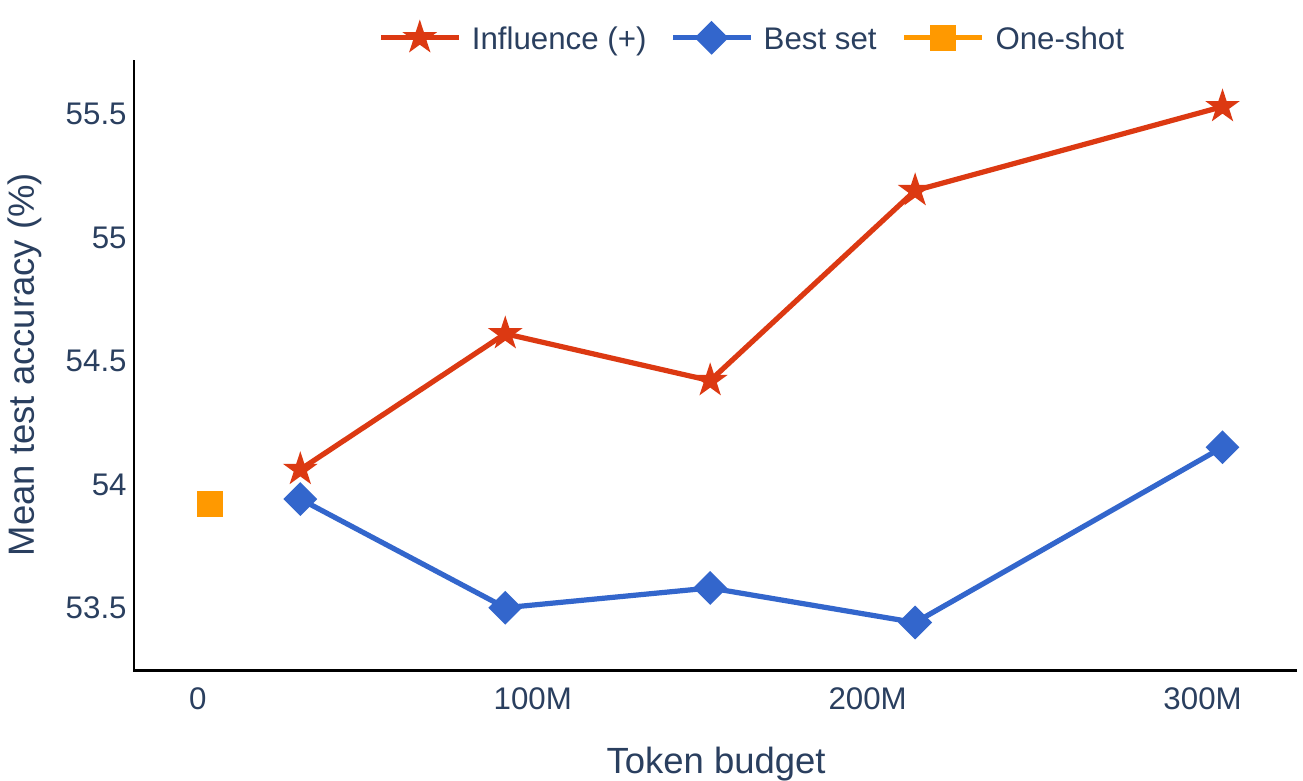}
    \vskip -0.05in
    \caption{Token budget comparison for different baselines evaluated on LLaMA-7B ($|S|=400$).}
    \label{fig_token_match}
\end{minipage}
\hfill
\begin{minipage}[t]{0.47\columnwidth}
    \includegraphics[width=\linewidth]{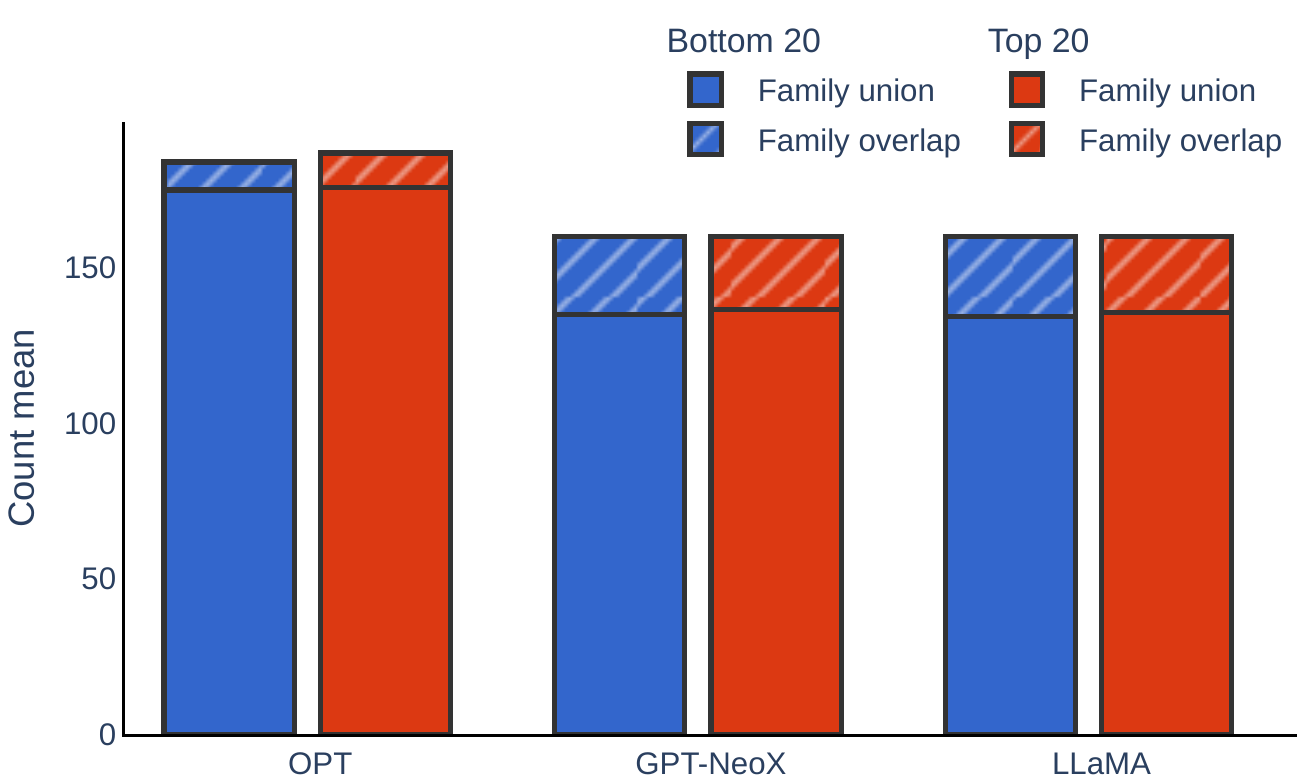}
    \vskip -0.05in
    \caption{Model family agreement (overlap) when considering all examples (union) in the \textcolor{Bittersweet}{Top} and \textcolor{Blue}{Bottom} $20^{th}$ influence bins.}
    \label{fig_family_agreement}
\end{minipage}
\end{figure}
This section analyzes whether or not the best and worst examples on a task are shared across models. When considering the overlap between all 7 individual models, our work finds that they rarely agree on the most positive and negative influence examples ($\leq4$ for all tasks). However, within the same model families, we identify a decent overlap. Figure~\ref{fig_family_agreement} plots the inter-family agreement between three families considered in our study. Compared to OPT, both GPT-NeoX and LLaMA models often identify a smaller set of top and bottom influence examples, while agreeing on these examples more often. This suggests that our in-context influences are picking up signals specific to the nature of these model families. These can include variations in model architecture, training data, training stability, tokenizers, and others \cite{zhang_opt_2022, black_gpt-neox-20b_2022, Radford2019LanguageMA,touvron2023llama}. For instance, model performance has been linked to term frequencies in the pretraining data \cite{termfrequency}.

\subsection{Negative vs. Positive Examples}
\begin{table}[t]
\caption{Negative examples identified by in-context influences on LLaMA-7B.}
\label{tab_negative-prompts}
\vskip 0.1in
\centering
\resizebox{\columnwidth}{!}{%
\begin{tabular}{@{}llllr@{}}
\toprule
~ & ID & Prompt & Influence & Reason \\ \midrule
PIQA & 12444 & \multirow{2}{*}{\begin{tabular}[c]{@{}l@{}}Goal: flashlight\\ Answer: shines a light\end{tabular}} & -0.001854 & Unnatural \\
 &  &  &  &  \\
WIC & 3890 & \multirow{5}{*}{\begin{tabular}[c]{@{}l@{}}Go to the supermarket and buy some tea.\\ Would you like some tea?\\ question: Is the word 'tea' used in the same sense in the \\ two sentences above?\\ answer: false\end{tabular}} & -0.007068 & Mislabeled \\
~ & ~ & ~ & ~ & \\
~ & ~ & ~ & ~ & \\
~ & ~ & ~ & ~ & \\
~ & ~ & ~ & ~ & \\
OBQA & 3771 & \multirow{2}{*}{\begin{tabular}[c]{@{}l@{}}Context: Single cell organisms can put an animal in the\\ Answer: emergency room\end{tabular}} & -0.006058 & Unnatural \\
 &  &  &  &  \\
\bottomrule
\end{tabular}
}
\end{table}
\begin{figure*}[t]
\label{fig_signals}
\centering
\centerline{\includegraphics[width=1.0\textwidth]{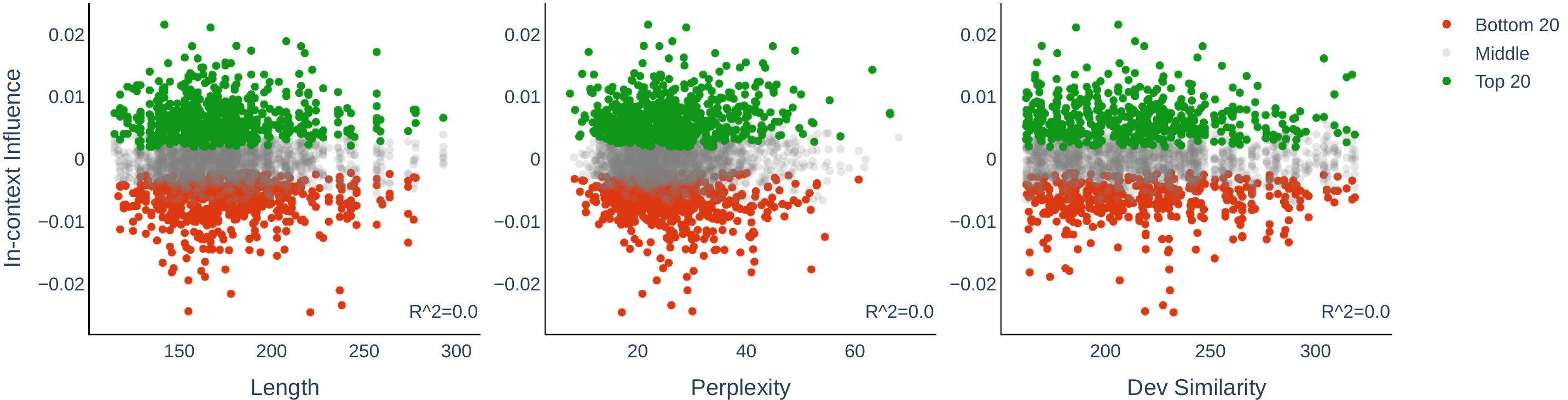}}
\caption{On Superglue-WIC, in-context influences do not correlate with any previously known example characteristics.}
\end{figure*}
Prior work has associated various characteristics with examples that are strongly positive or negative \cite{koh2017understanding, han_explaining_2020, ilyas_datamodels_2022}. Positive examples are found to sometimes be instances of data leakage during the training process, while negative examples are often mislabeled. For LLMs, we do not have access to the pretraining data to identify data leakage. However, we identify many instances in the bottom influence bin that appear as either ``unnatural'' or mislabeled. Table~\ref{tab_negative-prompts} shows instances of these negative examples and their associated potential cause. For PIQA example \texttt{\#12444}, the overall plausibility of the statement could improve if the order of statements \texttt{Goal} and \texttt{Answer} gets switched. Alternatively, a better template could possibly help achieve better input-output coherence. We suspect that the prompt template might play an important role in determining the influence of an example. Additionally, we identify WIC example \texttt{\#3890} as a falsely-annotated instance. Related to input-label mapping, \citet{min_rethinking_2022} has shown that label correctness is not important for good ICL performance.\footnote{See Table~\ref{tab_highprompts_llama} and Table~\ref{tab_lowprompts_llama} in the Appendix for more examples of highly positive and negative influence points.}

Quantitatively, we measure several metrics from the literature to compare examples with positive and negative influences.  
As Figure~\ref{fig_signals} illustrates, we find little to no association between in-context influences and known signals such as input length, perplexity, and similar distance to the input data ($R^2=0.0$) \cite{liu-etal-2022-makes,gonen_perplexity_2022}. This suggests that our influence-based selection framework has captured signals unrelated to these 3 metrics, and that model family is broadly a better predictor, as shown in Section~\ref{section:family-agreement}.

\paragraph{ICL Sensitivity.} Naturally, we can leverage in-context influences to quantify the gap between the most positive and negative examples in ICL. For example, on OpenBookQA, we observe an impressive $16.3\%$ accuracy difference between the best and worst in-context examples on LLaMA-13B (See Table~\ref{tab_influence-diff} in Appendix). Our framework adds to a list of previous works reporting ICL sensitivity \cite{liu-etal-2022-makes, lu-etal-2022-fantastically, pmlr-v139-zhao21c}.

\paragraph{Influence bins.} Additionally, we demonstrate that our influence framework can analyze example selection in more fine-grain. For this experiment, we group examples by their influence percentile, where each bin contains $20\%$ of the Train set. From these bins, we randomly select a set of $k$ examples in any ordering for 10 seeds for evaluation. If the influences are meaningful, we expect to see increasing performance gains as examples are selected in increasing percentile bins along the x-axis. Figure~\ref{fig_influence_bins} identifies clear positive trends confirming our hypothesis on most models and tasks. This shows that in-context influences produce well-behaved results when examples are selected in specific influence regions. There are few exceptions, such as the BoolQ task on LLaMA-7B, where selecting examples in increasingly positive influences does not consistently improve validation accuracy.

\subsection{How do in-context influences generalize across k-shot?}

\begin{figure*}[t]
\centering
\centerline{\includegraphics[width=1.0\textwidth]{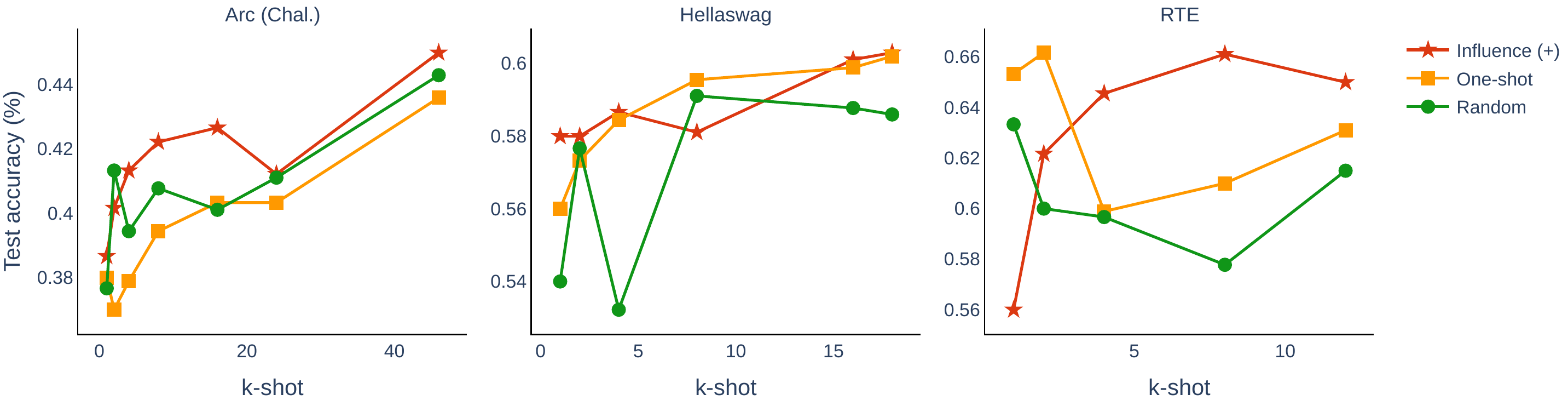}}
\caption{On LLaMA-7B, influence-based example selection scales well with increasing $k$-shot.}
\label{fig_scaling}
\end{figure*}

Thus far, our estimation of in-context influences has assumed a many-shot setting where a maximal number of examples is packed into the context window. In this study, we are interested in knowing how in-context influences generalize to different numbers of in-context examples $k$. In comparing different selection methods, Figure~\ref{fig_scaling} finds that the impact of in-context influences is most prevalent when $k$ is many (generally $\geq8$). At one-shot and very few-shot, in-context influences can sometimes perform worse than other methods (RTE) but steadily improve with increasing $k$. In contrast, the performance for One-shot and random selection do not always improve (and sometimes decline) with increasing $k$ (on Hellaswag and RTE).

\subsection{Case study: Example Ordering}
\label{section:case_study}
We want to apply our influence-based framework to demonstrate its effectiveness for studying a phenomenon in ICL, which deals with recency bias in example ordering \cite{lu-etal-2022-fantastically,pmlr-v139-zhao21c}.

\paragraph{Setup.} To do this, we randomly choose 100 examples from another SuperGLUE task, CB, and assign them into 4 groups for 4-shot prompting. On OPT-6.7B, we compute an in-context influence estimate for each example-position pair over all possible ordering permutations ($4! = 24$). Given an arbitrary example, we are interested in quantifying its impact at any position in the ordering and the overall influence of each position.

\begin{figure}[t]
\centering
\begin{minipage}[t]{0.47\columnwidth}
    \includegraphics[width=\linewidth]{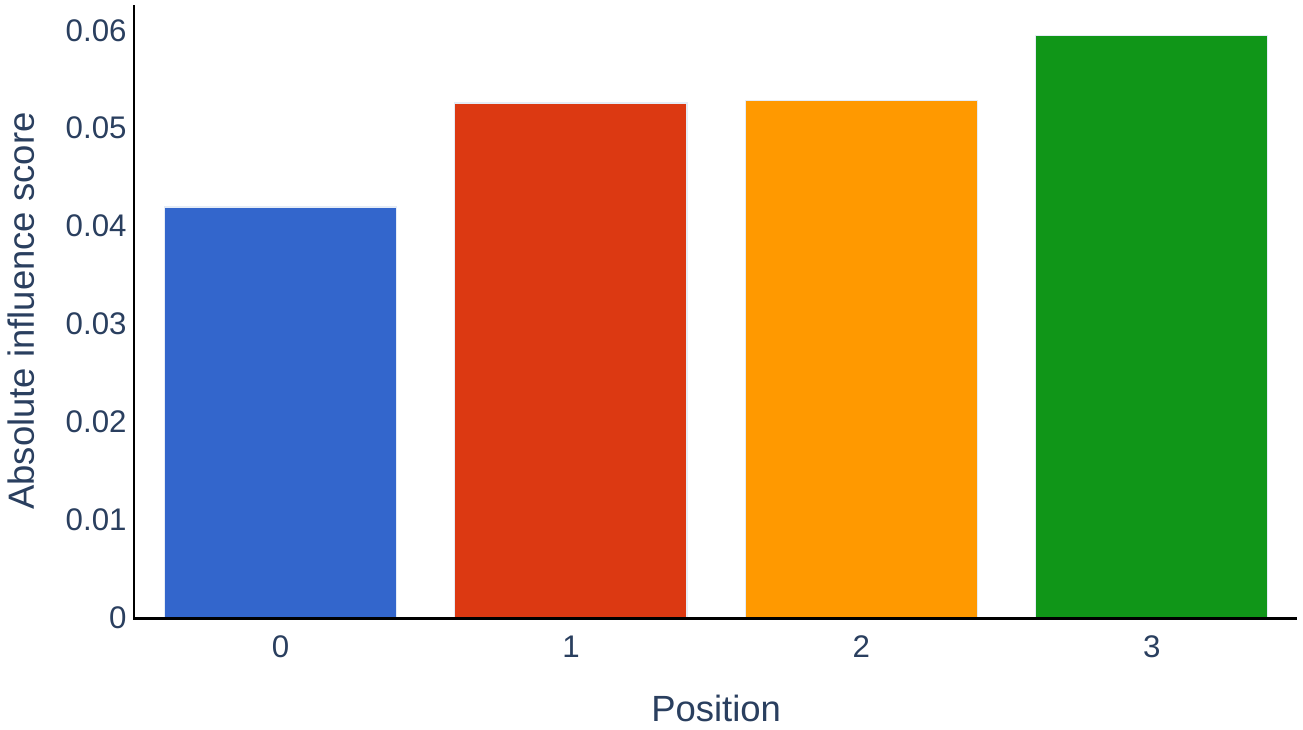}
    \caption{Aggregated influences of each position in 4-shot prompting. Influence magnitudes are bigger at later positions.}
    \label{fig_positions}
\end{minipage}
\hfill
\begin{minipage}[t]{0.47\columnwidth}
    \includegraphics[width=\linewidth]{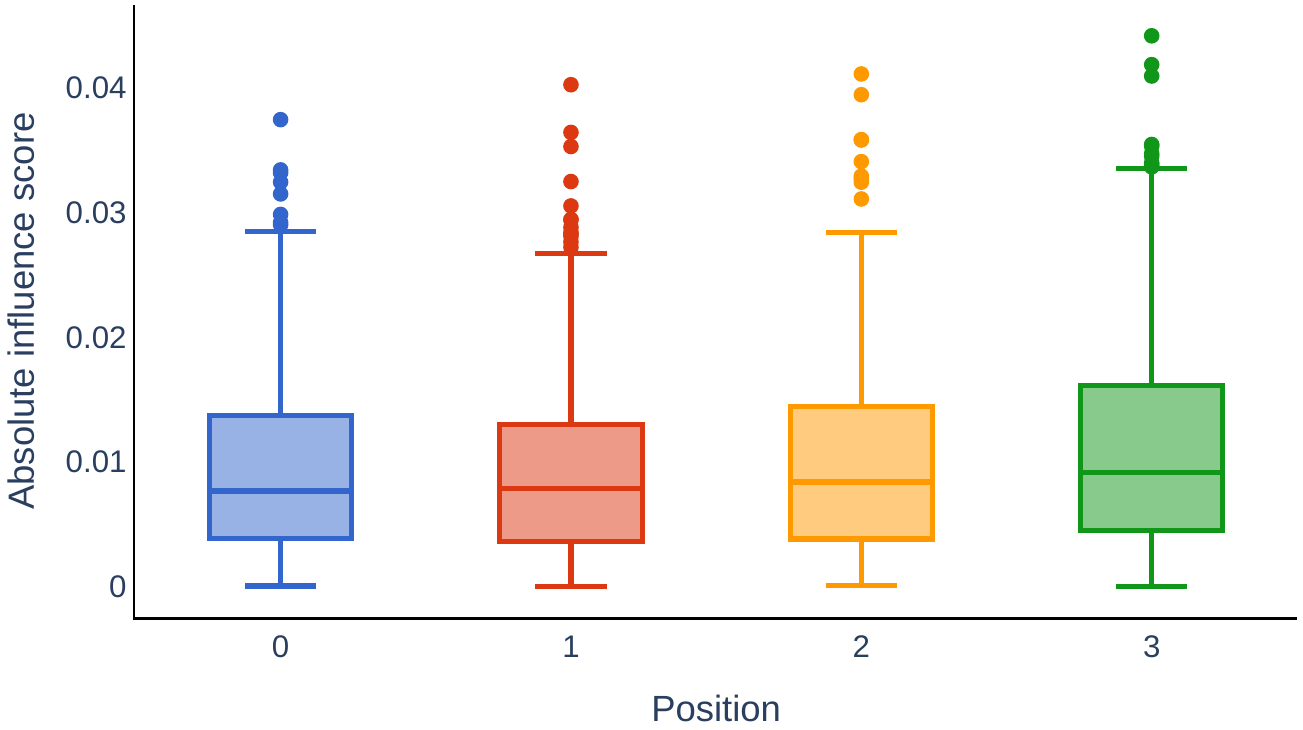}
    \caption{Influence distribution of each position in 4-shot prompting. Bigger spreads are observed at later positions.}
    \label{fig_positions-dist}
\end{minipage}
\end{figure}

\paragraph{Results.} Figure~\ref{fig_positions} confirms the presence of recency bias in ICL \cite{pmlr-v139-zhao21c}, showing that influence estimates of examples increase as their position ID moves down in the order. Between Position $\#0$ and Position $\#3$, there is a notable $2\%$ difference in the estimated absolute influence. Figure~\ref{fig_positions-dist} elaborates on this result: on the same set of in-context examples, the influence estimates computed in position $\#3$ has the biggest spread among all positions. Once again, we observe a steadily increasing trend in the widths of the spread as an example is moved down in order.
\section{Related Work}

\paragraph{Example selection.} In parallel and independent work, \citet{chang2022careful} also study the use of influences for selecting in-context examples for $k$-shot prompting, and also find that influence-based selection outperforms baseline methods. While we both consider influence estimates based on datamodels and data shapley influences, there are some differences. \citet{chang2022careful} integrate the position of an in-context example into the datamodel to directly calculate the influence of position for each example. In contrast, we consider the vanilla datamodel that does not model position, but demonstrate positional bias in a case study in Section~\ref{section:case_study}. Although the formulation of the \texttt{CondAcc} score from \citet{chang2022careful} may appear slightly different from our influence metric, \citet{chang2022careful} prove in their Appendix that the two quantities rank examples identically. The experimental setups cover two distinct use-cases -- \citet{chang2022careful} focus on a smaller number of in-context examples (i.e. $k=4$) and find that influences could greatly reduce the variance of ICL, while we study a large number of examples (i.e. $k$ up to $52$) that also leads to less variance and performance gains. Finally, the corresponding analyses complement each other well. \citet{chang2022careful} analyze the embedding distance of examples, while we analyze the scaling pattern of the number of in-context examples $k$ and the level of influence agreement across model families. Both work find little correlation between in-context influences and known signals such as example perplexity. The combined analyses present a more comprehensive understanding of influence-based example selection for ICL.

Outside of in-context influences, examples have been found to be inequal when used in ICL. \citet{liu-etal-2022-makes} finds that the best in-context examples are the ones most semantically similar to the test sample, which translates well for semantic parsing tasks. \citet{chen_sensitity_2022} links exemplars to a sensitivity measure, while \citet{gonen_perplexity_2022} recently shows a correlation between prompt perplexity and model performance. Others have improved ICL performance by focusing on prompt retrieval \cite{rubin}, or applying reinforcement learning to improve ICL performance through prompt editing \cite{zhang2022tempera}. Our in-context influence framework differs in its focus on identifying good examples from a Dev set that generalize well to any unseen evaluation, removing the need to perform any prompt editing or retrieval at test time.

\paragraph{In-context learning.} ICL comes with high volatility to factors beyond example selection. In the few-shot setting, models have shown a tendency to overly rely on the most frequent labels (majority bias) or labels that appear at late positions in a prompt (recency bias) \cite{pmlr-v139-zhao21c}. The latter suggests that the ordering of examples can be optimized for performance gain \cite{lu-etal-2022-fantastically}. The prompt template -- the format in which the example is presented -- also matters \cite{min-etal-2022-noisy}. Other findings have discovered that correct input-label mapping has little relevance \cite{min_rethinking_2022} and example diversity is more important \cite{su_selective}. Recently, \citet{learning-algo} links the underlying computations of ICL to linear algorithms.

\paragraph{Training data influence.} Influence functions \cite{koh2017understanding} have been used as a way to trace a model's output back to the training data. Influence of a specific training point measures the change in a model's performance when the point is removed from the training set. Data Shapley \cite{pmlr-ghorbani} and \citet{ilyas_datamodels_2022} measure similar quantities via retraining the model on subsets of the dataset.
%
Outside of individual attributions, influence functions have also been used to measure group effects, where prior work found the influence estimates of individual data points to be the lower bound of groups \cite{koh_groups_NEURIPS2019}.  
\section{Conclusion}
Our work proposes in-context influences as a way to analyze and select examples for ICL. Influence-based example selection methods (in-context influences and in-context datamodels) outperform all baselines for both positive and negative selections, showing stronger results on multi-choice tasks compared to binary classification tasks. In-context influences can identify problematic examples, scale performance with the choice of $k$-shot, and generalize to many nidek families. In a case study, we further examine known biases found in ICL such as recency bias in example ordering. Our work adds to a growing body of work that aims to understand and debug different emerging phenomena in LLMs. 

One limitation of influence-based frameworks is that they predict ICL performance from a fixed training set. However, practitioners can generate original prompts and examples, which may not exist in the training set. One potential research direction is to predict the performance of \textit{any} input example constructed on the fly, in addition to those in the training set. Our influence-based framework can also be leveraged to study ICL beyond classification performance. For example, future work can potentially calculate influences for other natural language tasks such as text generation, summarization, or other multi-task settings.
\newpage
\printbibliography

\newpage

\section{Full Results}
\label{appendix_fullresults}
We provide more results, experimental, and discussion details that did not fit into the main paper.

\subsection{Influence distribution}
\begin{table*}[h]
\caption{Mean difference of test accuracy ($\%$) between the top 20 and bottom 20 percentile bin for each model-task pair. The disparity between the two groups is clear, though it may vary by choice of model and task.}
\label{tab_influence-diff}
\vskip 0.1in
\begin{center}
\begin{tabular}{lrrrr}
\toprule
~ &  GPT-J &  OPT-6.7B &  LLaMA-7B &  LLaMA-13B \\
\midrule
PIQA                                  &   0.26 &     -1.47 &     -0.10 &       0.90 \\
BoolQ                                 &   0.20 &      1.33 &      4.30 &       4.40 \\
RTE                                   &   1.33 &      5.10 &     11.80 &      10.07 \\
WIC                                   &   4.23 &      2.00 &      3.57 &       3.00 \\
WSC                                   &  -5.84 &     -8.38 &     -1.69 &      10.77 \\
\midrule
Arc (Chal.) &  -0.83 &      1.16 &      0.66 &       4.03 \\
Arc (Easy)  &   0.23 &      0.00 &      2.70 &       0.96 \\
Hellaswag  &   2.10 &      1.54 &      3.04 &       3.32 \\
OBQA      &   4.27 &      5.96 &      7.60 &      16.34 \\
\bottomrule
\end{tabular}
\end{center}
\vskip -0.1in
\end{table*}


Table~\ref{tab_influence-diff} shows the performance gaps between using the most positive and the most negative influence examples for ICL inference.

\begin{figure*}[]
\label{fig_influence-full}
\vskip 0.2in
\begin{center}
\centerline{\includegraphics[width=1.0\textwidth]{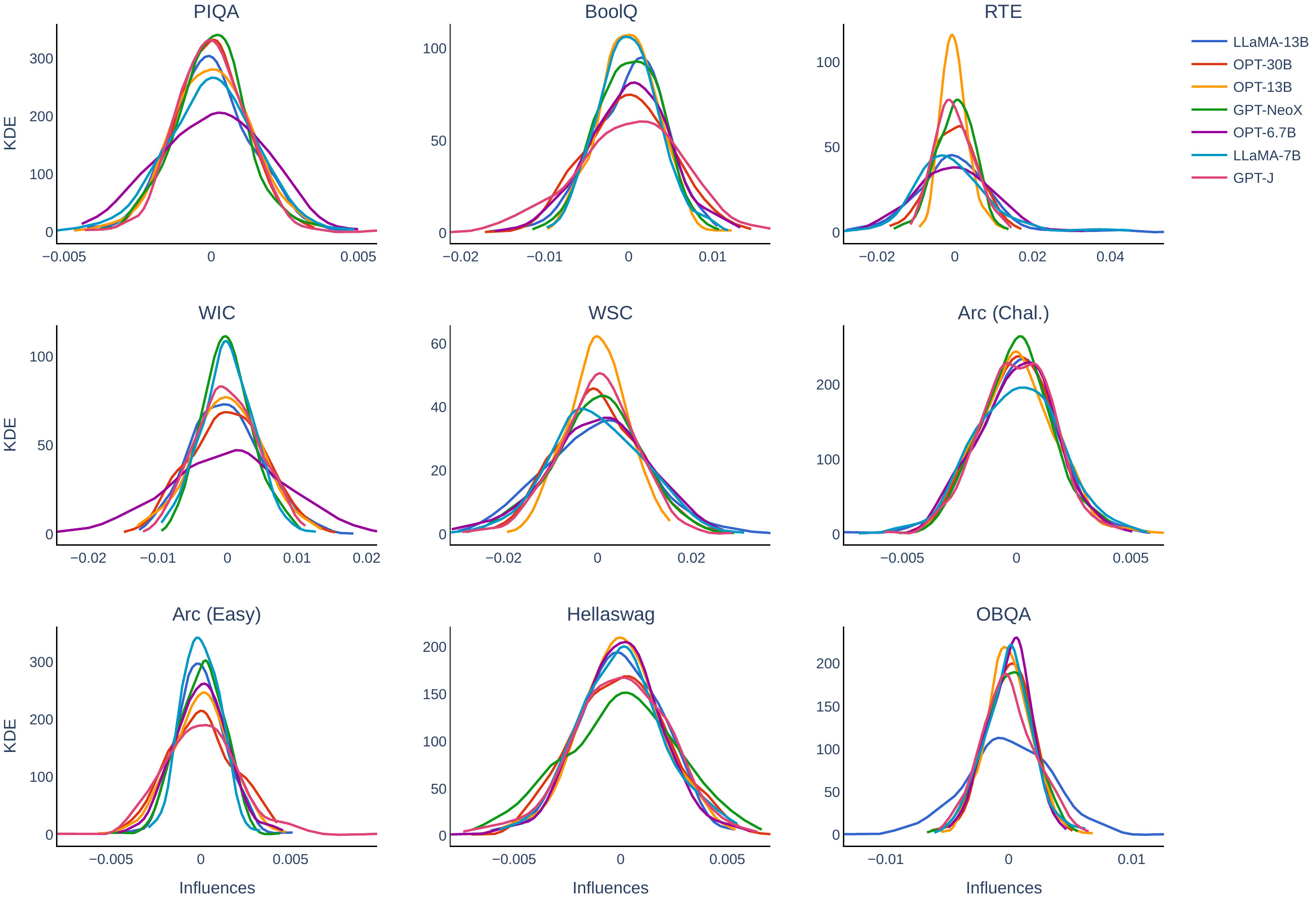}}
\end{center}
\vskip -0.2in
\caption{Influence distributions across all models and tasks. A wide spread signifies existence of many high-influence points.}
\end{figure*}

Figure~\ref{fig_influence-full} plots the distribution of influence estimates for all models and tasks.

\begin{figure*}[]
\label{fig_counterfactual-full}
\vskip 0.2in
\begin{center}
\centerline{\includegraphics[width=1.0\textwidth]{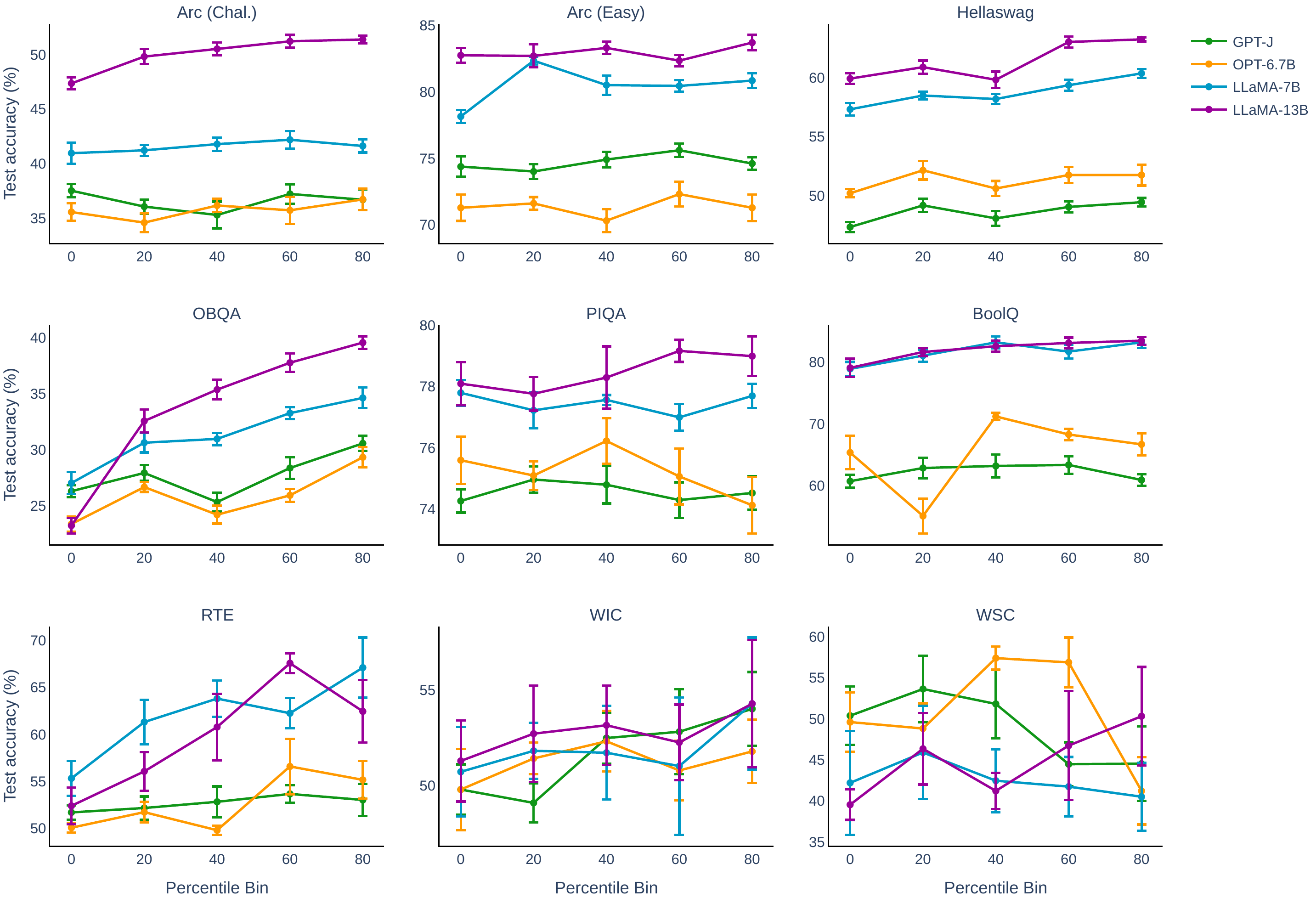}}
\end{center}
\vskip -0.2in
\caption{In most models and tasks, Test accuracy increases when in-context examples are selected in increasing influence percentile bins. Many task and model observes linear trends outside of few exceptions (ie. WSC).}
\end{figure*}

Figure~\ref{fig_counterfactual-full} visualizes the fine-grain behavior of influence-based example selection when examples are selected in increasing order of influences.

\subsection{Choice of $k$-shot for in-context influences}

\begin{figure*}[]
\label{fig_scaling-shot-full}
\vskip 0.2in
\begin{center}
\centerline{\includegraphics[width=1.0\textwidth]{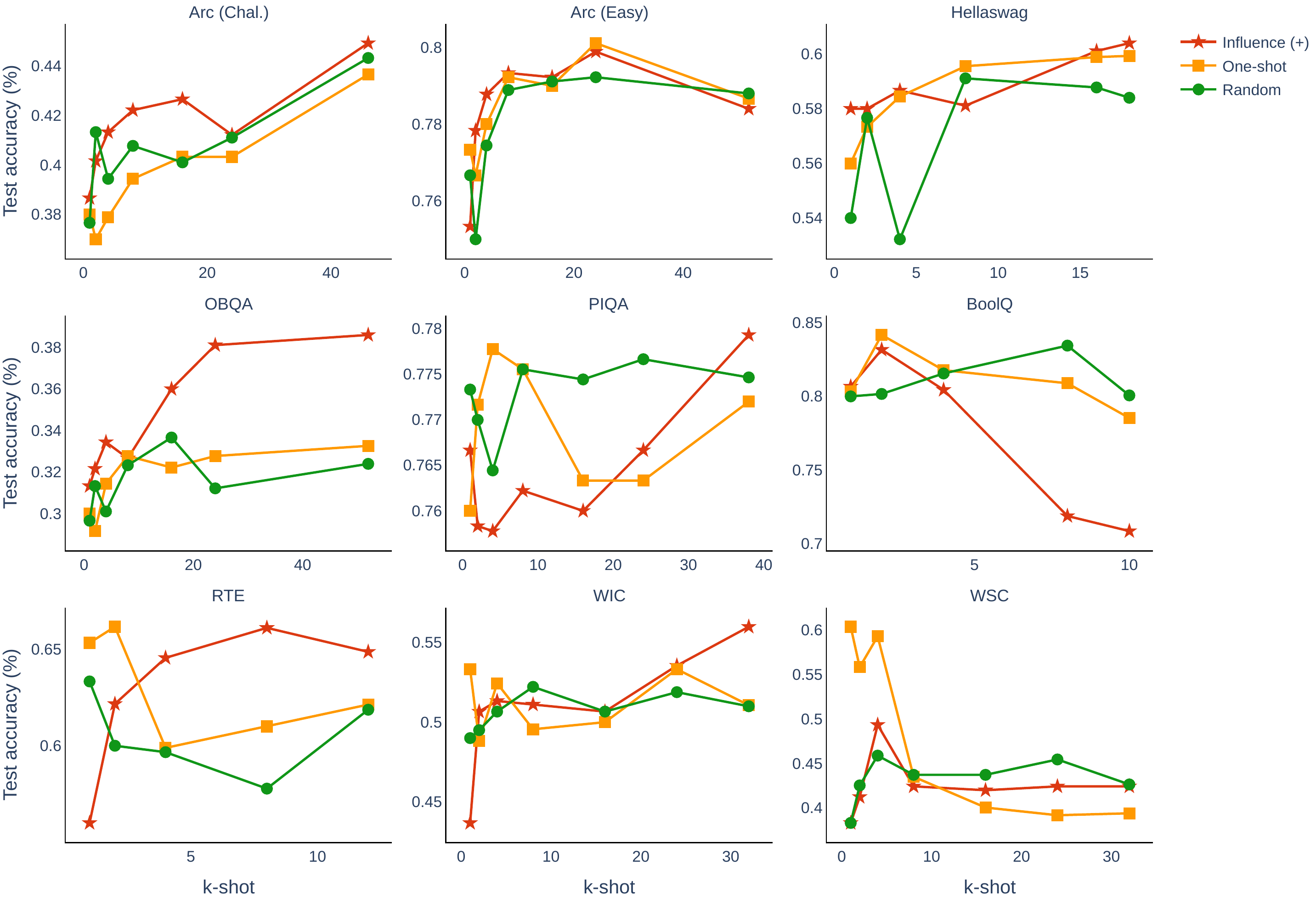}}
\end{center}
\vskip -0.2in
\caption{How different positive example selection methods generalize with the number of $k$ demonstrations.}
\end{figure*}

Figure~\ref{fig_scaling-shot-full} compares different example selection methods as $k$-shot scales.


\section{Discussion}
\subsection{Can linear datamodels predict in-context learning?}
\label{appendix_datamodels}

Recall from Section~\ref{section:infl-baselines} that we train linear in-context datamodels to derive $\theta$ as another influence measure for example selection. 
To evaluate these datamodels, we hold out a fraction of the training pairs (arbitrarily selected) from the collection process and use them afterwards as the ground truth. We do this for each model and task combination. If the Pearson correlation ($\rho$) between the predicted outputs and actual outputs are strong and statistically significant, we say that the linear datamodels models have capably captured the relationship between the in-context examples and ICL performance.

\begin{figure*}[]
\label{fig_datamodels-full}
\vskip 0.2in
\begin{center}
\centerline{\includegraphics[width=1.0\textwidth]{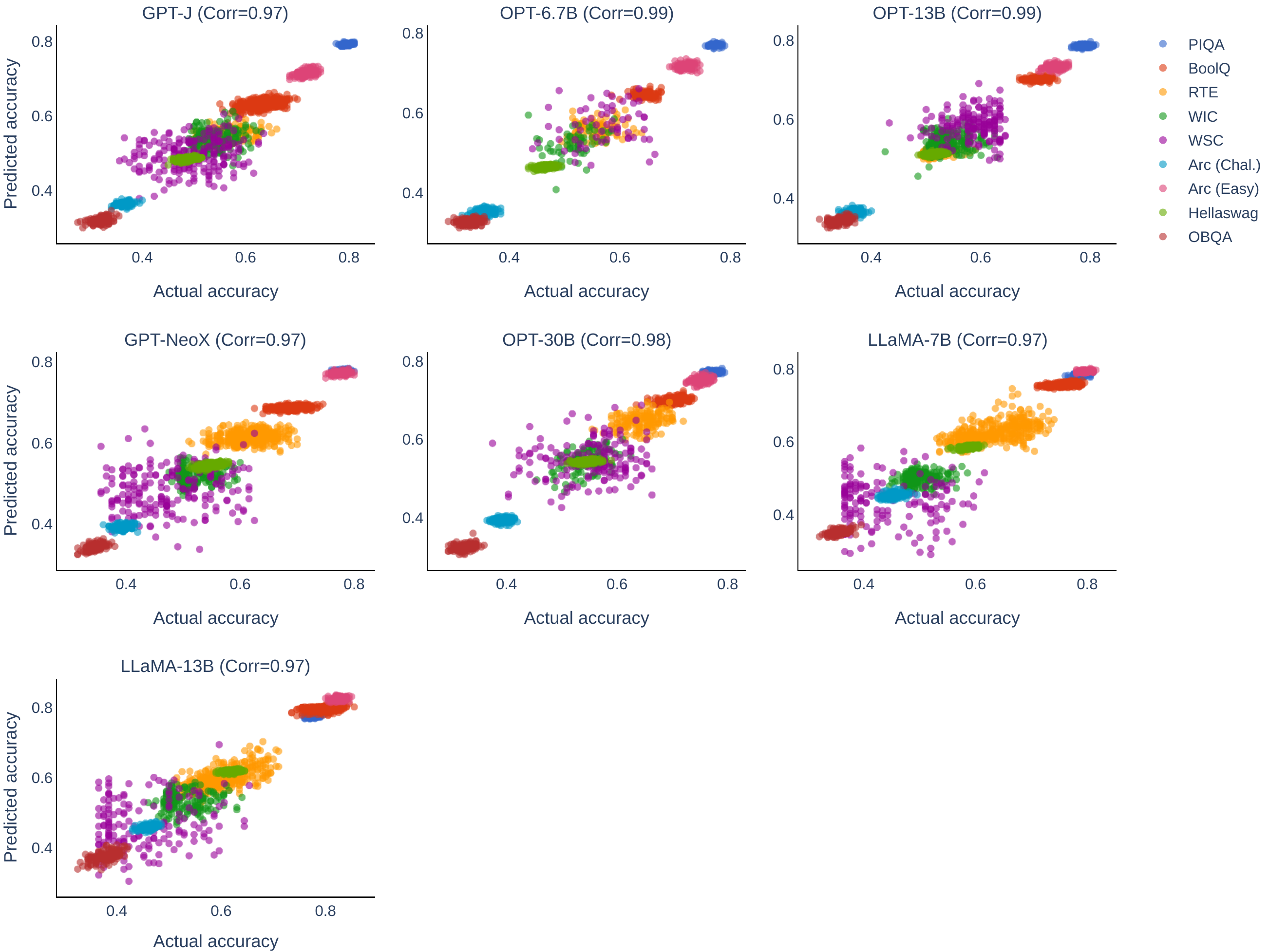}}
\end{center}
\vskip -0.2in
\caption{Linear in-context datamodels \textit{can} predict ICL performance on arbitrary subset $S'$. Pearson correlation is calculated over all tasks for the model.}
\end{figure*}

Figure~\ref{fig_datamodels-full} visualizes the correlation between the predicted and actual model outputs for all models and tasks. We observe strong linear trends across the board, implying that the fitted in-context datamodels can predict few-shot ICL performance on unseen subsets of examples. Among all tasks, SuperGLUE-WSC is the most difficult to predict, which can be explained by the high variance from having the smallest Test set.


\subsection{Erratic behavior with OPT models on SuperGLUE}
\label{appendix_optoddity}
Authors of OPT report the model's erratic behaviors when evaluated on many SuperGLUE tasks \cite{zhang_opt_2022}. Specifically, on the task WSC, zero- and multi-shot performance do not improve with respect to scale. They suspect that the small size of the validation sets in these datasets can be a factor. There are also reported accidents during the training process related to hardware failures and loss divergences. These factors could partially explain signals found in our in-context influence estimates.


\section{Implementation Details}

\subsection{Models}
\label{appendix_model}

\paragraph{Language models.} All autoregressive models are downloaded from their HuggingFace checkpoints using the \texttt{transformers} module\footnote{\url{https://github.com/huggingface/transformers}}. To conserve memory, we load all models in \texttt{16FP} half precision. We thank the authors of these models for making their work available to the research community.

\paragraph{Sentences encoders.} For the Similarity baseline, we use RoBERTa-large\footnote{\url{https://huggingface.co/sentence-transformers/all-roberta-large-v1}} \cite{roberta}
sentence encoder provided by Huggingface's \texttt{sentence-transformer} module.

\paragraph{Seed.} By default, we keep a fixed \texttt{seed=42}. For experiments involving random example ordering, we also use other seeds in \texttt{\{51, 56, 67, 75, 82, 98\}}.

\subsection{Datasets}
\label{appendix_dataset}
All datasets were downloaded using Huggingface's \texttt{datasets} module. Table~\ref{tab_datasets} details the sizes of the subsampled sets and the number of shots that fit in the in-context windows.
\begin{itemize}
    \item \textbf{SuperGLUE \cite{superglue}} This benchmark includes 5 binary classification tasks: BoolQ, RTE, WIC, WSC, and CB.
    \item \textbf{PIQA \cite{ai2-arc}} This benchmark includes 2 multi-choice tasks: AI2 Arc (Challenge) \& AI2 Arc (Easy).
    \item \textbf{Hellaswag \cite{hellaswag}} Single multi-choice task: Hellaswag.
    \item \textbf{OpenBookQA \cite{openbookqa}} Single multi-choice task: OpenBookQA.
\end{itemize}

\begin{table}[ht]
\caption{Models used in our work.}
\label{tab_models}
\vskip 0.1in
\centering
\begin{tabular}{lccr}
\toprule
Model & Parameter \# & Window & Open-source\\
\midrule
GPT-J & 6B & 2048 & $\surd$\\
GPT-NeoX & 20B & 2048 & $\surd$\\
OPT-6.7B & 6.7B & 2048 & $\surd$\\
OPT-13B & 13B & 2048 & $\surd$\\
OPT-30B & 30B & 2048 & $\surd$ \\
LLaMA-7B & 7B & 2048 & $\surd$\\
LLaMA-13B & 13B & 2048 & $\surd$ \\
\bottomrule
\end{tabular}
\vskip -0.1in
\end{table}
\begin{table}[ht]
\caption{
Datasets used in the paper. We sample 400 examples for train and 200 examples for validation wherever possible. $k$ denotes the number of demonstrations necessary to fill up the 2048 character limit of context windows.}
\label{tab_datasets}
\begin{center}
\begin{tabular}{lccccr}
\toprule
~ & Type & $\mid$ Train $\mid$ & $\mid$ Dev $\mid$ & $\mid$ Test $\mid$ & $k$\\
\midrule
PIQA & Binary & 400 & 200 & 500 & 38\\
Superglue BoolQ & Binary & 400 & 200  & 500 & 10\\
Superglue RTE & Binary & 400 & 200  & 500 & 12\\
Superglue WIC & Binary & 400 & 200  & 500 & 32\\
Superglue WSC & Binary & 400 & 104  & 154 & 32\\
\midrule
AI2 Arc (Challenge) & MC & 400 & 200  & 500 & 46\\
AI2 Arc (Easy) & MC & 400 & 200  & 500 & 52\\
Hellaswag & MC & 400 & 200  & 500 & 18\\
OpenBookQA & MC & 400 & 200  & 500 & 52\\
\bottomrule
\end{tabular}
\end{center}
\vskip -0.1in
\end{table}

\subsection{Prompts}
\label{appendix_prompt}
Table~\ref{tab_prompts} shows full prompt formats used the paper.

\subsection{Hardware}
We run all experiments on the \texttt{NVIDIA A100} and \texttt{NVIDIA RTX A6000} GPUs.
\clearpage
\newpage

\clearpage
\newpage
\setlength{\LTcapwidth}{\textwidth}
\begin{longtable}{llllllllllc}
\caption{Full results for positive selection across all models over 7 seeds.}
\label{tab_mainresults-pos-full} \\
\toprule
{} &                               PIQA &                            BoolQ &                                RTE &                               WIC &                              WSC & ARC-c & ARC-e & HS &                             OBQA &   Rank ($\downarrow$)\\
\midrule
\textbf{GPT-J-6B} & ~ & ~ & ~ & ~ & ~ & ~ & ~ & ~ & ~ & ~ \\
\midrule
One-shot (+)      &           75.0\scalebox{0.5}{0.0} &           53.3\scalebox{0.5}{0.1} &           50.0\scalebox{0.5}{0.0} &           50.0\scalebox{0.5}{0.0} &  \textbf{61.7\scalebox{0.5}{0.0}} &           37.6\scalebox{0.5}{0.0} &           73.7\scalebox{0.5}{0.0} &           49.1\scalebox{0.5}{0.0} &           30.6\scalebox{0.5}{0.0} &  4.57 \\
Random            &           75.7\scalebox{0.5}{0.1} &           61.7\scalebox{0.5}{0.3} &           56.2\scalebox{0.5}{0.3} &           51.9\scalebox{0.5}{0.3} &           48.5\scalebox{0.5}{0.4} &           37.7\scalebox{0.5}{0.1} &           73.3\scalebox{0.5}{0.0} &           49.1\scalebox{0.5}{0.1} &           27.6\scalebox{0.5}{0.1} &  4.32 \\
Perplexity (+)    &  \textbf{76.2\scalebox{0.5}{0.0}} &           64.4\scalebox{0.5}{0.1} &           54.1\scalebox{0.5}{0.2} &           50.1\scalebox{0.5}{0.0} &           38.5\scalebox{0.5}{0.2} &           38.4\scalebox{0.5}{0.1} &           73.1\scalebox{0.5}{0.0} &           49.1\scalebox{0.5}{0.0} &           26.8\scalebox{0.5}{0.0} &  4.22 \\
Similarity (+)    &           75.5\scalebox{0.5}{0.0} &           61.8\scalebox{0.5}{0.1} &           59.0\scalebox{0.5}{0.2} &           51.9\scalebox{0.5}{0.2} &           55.8\scalebox{0.5}{0.4} &           37.6\scalebox{0.5}{0.1} &           72.9\scalebox{0.5}{0.0} &           49.7\scalebox{0.5}{0.0} &           27.0\scalebox{0.5}{0.0} &  4.06 \\
IC Datamodels (+) &           75.3\scalebox{0.5}{0.0} &           57.5\scalebox{0.5}{0.1} &           50.5\scalebox{0.5}{0.0} &  \textbf{55.7\scalebox{0.5}{0.2}} &           46.0\scalebox{0.5}{0.3} &           38.5\scalebox{0.5}{0.1} &           73.6\scalebox{0.5}{0.0} &           50.5\scalebox{0.5}{0.0} &  \textbf{30.9\scalebox{0.5}{0.0}} &  3.46 \\
Influence (+)     &           75.4\scalebox{0.5}{0.0} &           62.1\scalebox{0.5}{0.2} &           50.4\scalebox{0.5}{0.1} &           55.3\scalebox{0.5}{0.2} &           49.7\scalebox{0.5}{0.6} &  \textbf{39.4\scalebox{0.5}{0.1}} &           73.5\scalebox{0.5}{0.1} &  \textbf{51.1\scalebox{0.5}{0.0}} &           30.5\scalebox{0.5}{0.1} &  3.17 \\
Best set          &           76.0\scalebox{0.5}{0.0} &  \textbf{65.0\scalebox{0.5}{0.1}} &  \textbf{59.2\scalebox{0.5}{0.2}} &           51.7\scalebox{0.5}{0.2} &           52.7\scalebox{0.5}{0.3} &           37.7\scalebox{0.5}{0.1} &  \textbf{73.9\scalebox{0.5}{0.1}} &           49.4\scalebox{0.5}{0.0} &           29.6\scalebox{0.5}{0.0} &  \textbf{3.14} \\
\bottomrule

~ & ~ & ~ & ~ & ~ & ~ & ~ & ~ & ~ & ~ & ~ \\
\textbf{GPT-NeoX-20B} & ~ & ~ & ~ & ~ & ~ & ~ & ~ & ~ & ~ & ~ \\
\midrule
Perplexity (+)    &           76.6\scalebox{0.5}{0.0} &           73.2\scalebox{0.5}{0.1} &           63.2\scalebox{0.5}{0.2} &  \textbf{51.9\scalebox{0.5}{0.2}} &           42.3\scalebox{0.5}{0.4} &           43.2\scalebox{0.5}{0.0} &           78.0\scalebox{0.5}{0.0} &           54.4\scalebox{0.5}{0.0} &           29.5\scalebox{0.5}{0.0} &  4.97 \\
Random            &           77.3\scalebox{0.5}{0.0} &           67.0\scalebox{0.5}{0.4} &           62.4\scalebox{0.5}{0.3} &           51.7\scalebox{0.5}{0.2} &           43.5\scalebox{0.5}{0.6} &           43.9\scalebox{0.5}{0.1} &           77.8\scalebox{0.5}{0.1} &           55.1\scalebox{0.5}{0.1} &           30.3\scalebox{0.5}{0.0} &  4.41 \\
Similarity (+)    &           77.0\scalebox{0.5}{0.0} &           64.1\scalebox{0.5}{0.5} &           65.0\scalebox{0.5}{0.3} &           51.2\scalebox{0.5}{0.1} &           56.0\scalebox{0.5}{0.6} &           43.5\scalebox{0.5}{0.1} &           77.9\scalebox{0.5}{0.1} &           54.8\scalebox{0.5}{0.1} &           29.8\scalebox{0.5}{0.1} &  4.35 \\
Best set          &           77.6\scalebox{0.5}{0.0} &           73.7\scalebox{0.5}{0.1} &           63.7\scalebox{0.5}{0.2} &           50.5\scalebox{0.5}{0.2} &           46.3\scalebox{0.5}{0.5} &           43.5\scalebox{0.5}{0.0} &           77.6\scalebox{0.5}{0.1} &           55.1\scalebox{0.5}{0.1} &           31.3\scalebox{0.5}{0.1} &  4.08 \\
One-shot (+)      &           76.5\scalebox{0.5}{0.0} &           57.3\scalebox{0.5}{0.1} &           53.5\scalebox{0.5}{0.2} &           48.9\scalebox{0.5}{0.1} &  \textbf{61.7\scalebox{0.5}{0.0}} &  \textbf{44.5\scalebox{0.5}{0.1}} &           78.5\scalebox{0.5}{0.1} &  \textbf{55.9\scalebox{0.5}{0.0}} &           32.8\scalebox{0.5}{0.0} &  4.06 \\
Influence (+)     &  \textbf{78.0\scalebox{0.5}{0.0}} &           73.6\scalebox{0.5}{0.2} &           65.3\scalebox{0.5}{0.1} &  \textbf{51.9\scalebox{0.5}{0.2}} &           47.2\scalebox{0.5}{0.4} &           43.8\scalebox{0.5}{0.1} &           78.7\scalebox{0.5}{0.1} &           54.9\scalebox{0.5}{0.1} &           32.7\scalebox{0.5}{0.0} &  2.95 \\
IC Datamodels (+) &           77.7\scalebox{0.5}{0.0} &  \textbf{75.2\scalebox{0.5}{0.0}} &  \textbf{66.3\scalebox{0.5}{0.2}} &           51.6\scalebox{0.5}{0.2} &           44.3\scalebox{0.5}{0.4} &           44.1\scalebox{0.5}{0.0} &  \textbf{79.5\scalebox{0.5}{0.1}} &           55.5\scalebox{0.5}{0.1} &  \textbf{32.9\scalebox{0.5}{0.0}} &  \textbf{2.52} \\
\bottomrule

~ & ~ & ~ & ~ & ~ & ~ & ~ & ~ & ~ & ~ & ~ \\
\textbf{LLaMA-7B} & ~ & ~ & ~ & ~ & ~ & ~ & ~ & ~ & ~ & ~ \\
\midrule
Similarity (+)    &           77.8\scalebox{0.5}{0.0} &           79.6\scalebox{0.5}{0.1} &           64.3\scalebox{0.5}{0.3} &           52.5\scalebox{0.5}{0.3} &           40.2\scalebox{0.5}{0.3} &           44.7\scalebox{0.5}{0.1} &           76.9\scalebox{0.5}{0.1} &           58.2\scalebox{0.5}{0.0} &           31.9\scalebox{0.5}{0.1} &  4.60 \\
One-shot (+)      &           77.2\scalebox{0.5}{0.0} &           78.2\scalebox{0.5}{0.1} &           60.9\scalebox{0.5}{0.3} &           51.6\scalebox{0.5}{0.1} &           39.1\scalebox{0.5}{0.1} &           43.6\scalebox{0.5}{0.0} &           78.7\scalebox{0.5}{0.0} &           59.9\scalebox{0.5}{0.0} &           33.5\scalebox{0.5}{0.1} &  4.52 \\
Perplexity (+)    &  \textbf{78.2\scalebox{0.5}{0.0}} &           78.1\scalebox{0.5}{0.1} &  \textbf{68.1\scalebox{0.5}{0.1}} &           50.4\scalebox{0.5}{0.1} &           42.4\scalebox{0.5}{0.6} &           44.2\scalebox{0.5}{0.1} &           77.3\scalebox{0.5}{0.1} &           59.3\scalebox{0.5}{0.0} &           32.0\scalebox{0.5}{0.1} &  4.32 \\
Random            &           78.0\scalebox{0.5}{0.1} &           80.3\scalebox{0.5}{0.1} &           63.4\scalebox{0.5}{0.2} &           51.1\scalebox{0.5}{0.2} &           41.7\scalebox{0.5}{0.3} &           44.3\scalebox{0.5}{0.0} &           78.7\scalebox{0.5}{0.1} &           58.8\scalebox{0.5}{0.1} &           32.3\scalebox{0.5}{0.0} &  4.11 \\
Influence (+)     &           78.0\scalebox{0.5}{0.0} &           72.7\scalebox{0.5}{0.3} &           65.7\scalebox{0.5}{0.2} &  \textbf{54.0\scalebox{0.5}{0.3}} &           43.0\scalebox{0.5}{0.4} &           44.8\scalebox{0.5}{0.0} &           78.7\scalebox{0.5}{0.1} &           60.0\scalebox{0.5}{0.0} &           38.3\scalebox{0.5}{0.0} &  3.25 \\
IC Datamodels (+) &           77.7\scalebox{0.5}{0.0} &           75.4\scalebox{0.5}{0.3} &           65.1\scalebox{0.5}{0.1} &           52.9\scalebox{0.5}{0.3} &           42.3\scalebox{0.5}{0.4} &  \textbf{45.4\scalebox{0.5}{0.0}} &  \textbf{78.9\scalebox{0.5}{0.0}} &  \textbf{60.2\scalebox{0.5}{0.1}} &  \textbf{38.7\scalebox{0.5}{0.1}} &  3.19 \\
Best set          &  \textbf{78.2\scalebox{0.5}{0.0}} &  \textbf{81.1\scalebox{0.5}{0.1}} &           67.6\scalebox{0.5}{0.1} &           51.7\scalebox{0.5}{0.1} &  \textbf{44.5\scalebox{0.5}{0.3}} &           45.0\scalebox{0.5}{0.1} &           77.7\scalebox{0.5}{0.0} &           59.8\scalebox{0.5}{0.0} &           35.1\scalebox{0.5}{0.0} &  \textbf{2.90} \\
\bottomrule

~ & ~ & ~ & ~ & ~ & ~ & ~ & ~ & ~ & ~ & ~ \\
\textbf{LLaMA-13B} & ~ & ~ & ~ & ~ & ~ & ~ & ~ & ~ & ~ & ~ \\
\midrule
Similarity (+)    &           78.5\scalebox{0.5}{0.0} &           81.9\scalebox{0.5}{0.1} &           57.3\scalebox{0.5}{0.3} &           54.0\scalebox{0.5}{0.3} &           42.2\scalebox{0.5}{0.6} &           50.1\scalebox{0.5}{0.1} &           82.8\scalebox{0.5}{0.0} &           62.1\scalebox{0.5}{0.0} &           36.3\scalebox{0.5}{0.1} &  4.78 \\
Perplexity (+)    &  \textbf{79.1\scalebox{0.5}{0.0}} &           82.4\scalebox{0.5}{0.1} &           61.8\scalebox{0.5}{0.3} &  \textbf{55.0\scalebox{0.5}{0.2}} &           40.8\scalebox{0.5}{0.1} &           50.5\scalebox{0.5}{0.0} &           82.5\scalebox{0.5}{0.0} &           61.6\scalebox{0.5}{0.0} &           35.6\scalebox{0.5}{0.1} &  4.57 \\
Random            &           78.5\scalebox{0.5}{0.1} &           82.6\scalebox{0.5}{0.1} &           61.1\scalebox{0.5}{0.3} &           51.8\scalebox{0.5}{0.2} &           42.9\scalebox{0.5}{0.4} &           50.4\scalebox{0.5}{0.1} &           82.7\scalebox{0.5}{0.0} &           62.5\scalebox{0.5}{0.1} &           35.7\scalebox{0.5}{0.1} &  4.51 \\
One-shot (+)      &           78.1\scalebox{0.5}{0.1} &  \textbf{84.5\scalebox{0.5}{0.1}} &           58.3\scalebox{0.5}{0.1} &           50.0\scalebox{0.5}{0.0} &           38.3\scalebox{0.5}{0.0} &           52.6\scalebox{0.5}{0.0} &           82.3\scalebox{0.5}{0.0} &           63.3\scalebox{0.5}{0.0} &           38.8\scalebox{0.5}{0.1} &  4.43 \\
Best set          &           78.7\scalebox{0.5}{0.0} &           83.2\scalebox{0.5}{0.1} &  \textbf{69.4\scalebox{0.5}{0.3}} &           54.7\scalebox{0.5}{0.2} &  \textbf{46.9\scalebox{0.5}{0.5}} &           51.9\scalebox{0.5}{0.1} &           82.5\scalebox{0.5}{0.0} &  \textbf{63.7\scalebox{0.5}{0.0}} &           36.4\scalebox{0.5}{0.1} &  3.14 \\
IC Datamodels (+) &           78.8\scalebox{0.5}{0.0} &           83.9\scalebox{0.5}{0.1} &           66.6\scalebox{0.5}{0.1} &           54.2\scalebox{0.5}{0.2} &           41.9\scalebox{0.5}{0.5} &           52.9\scalebox{0.5}{0.0} &  \textbf{82.9\scalebox{0.5}{0.0}} &           62.3\scalebox{0.5}{0.0} &           42.5\scalebox{0.5}{0.0} &  2.86 \\
Influence (+)     &           78.7\scalebox{0.5}{0.0} &           84.3\scalebox{0.5}{0.1} &           68.4\scalebox{0.5}{0.2} &           54.5\scalebox{0.5}{0.1} &           42.5\scalebox{0.5}{0.5} &  \textbf{53.1\scalebox{0.5}{0.0}} &           82.7\scalebox{0.5}{0.0} &           62.6\scalebox{0.5}{0.0} &  \textbf{42.7\scalebox{0.5}{0.1}} &  \textbf{2.70} \\
\bottomrule

~ & ~ & ~ & ~ & ~ & ~ & ~ & ~ & ~ & ~ & ~ \\
\textbf{OPT-6.7B} & ~ & ~ & ~ & ~ & ~ & ~ & ~ & ~ & ~ & ~ \\
\midrule
Perplexity (+)    &           76.0\scalebox{0.5}{0.0} &  \textbf{69.8\scalebox{0.5}{0.1}} &           51.1\scalebox{0.5}{0.0} &           49.3\scalebox{0.5}{0.1} &           47.6\scalebox{0.5}{0.6} &           37.6\scalebox{0.5}{0.1} &           69.6\scalebox{0.5}{0.0} &           53.2\scalebox{0.5}{0.0} &           25.4\scalebox{0.5}{0.1} &  4.78 \\
Similarity (+)    &           75.5\scalebox{0.5}{0.0} &           66.9\scalebox{0.5}{0.2} &           53.6\scalebox{0.5}{0.3} &           50.8\scalebox{0.5}{0.1} &           58.3\scalebox{0.5}{0.3} &  \textbf{39.0\scalebox{0.5}{0.1}} &           69.9\scalebox{0.5}{0.1} &           52.1\scalebox{0.5}{0.1} &           26.9\scalebox{0.5}{0.1} &  4.38 \\
Random            &           75.5\scalebox{0.5}{0.1} &           68.2\scalebox{0.5}{0.3} &           55.4\scalebox{0.5}{0.3} &           51.7\scalebox{0.5}{0.2} &           49.4\scalebox{0.5}{0.5} &           38.3\scalebox{0.5}{0.1} &           70.4\scalebox{0.5}{0.1} &           52.0\scalebox{0.5}{0.1} &           27.7\scalebox{0.5}{0.1} &  4.25 \\
One-shot (+)      &  \textbf{76.2\scalebox{0.5}{0.0}} &           58.1\scalebox{0.5}{0.1} &           55.4\scalebox{0.5}{0.3} &           50.0\scalebox{0.5}{0.0} &  \textbf{61.7\scalebox{0.5}{0.0}} &           38.1\scalebox{0.5}{0.1} &           68.9\scalebox{0.5}{0.1} &           53.0\scalebox{0.5}{0.1} &           30.5\scalebox{0.5}{0.1} &  3.97 \\
Best set          &           75.3\scalebox{0.5}{0.0} &           69.1\scalebox{0.5}{0.1} &           57.5\scalebox{0.5}{0.3} &           50.7\scalebox{0.5}{0.1} &           50.9\scalebox{0.5}{0.3} &           38.3\scalebox{0.5}{0.1} &           70.9\scalebox{0.5}{0.1} &           52.6\scalebox{0.5}{0.0} &           27.9\scalebox{0.5}{0.0} &  3.89 \\
IC Datamodels (+) &           75.6\scalebox{0.5}{0.0} &           68.5\scalebox{0.5}{0.1} &           59.6\scalebox{0.5}{0.1} &  \textbf{55.0\scalebox{0.5}{0.2}} &           53.2\scalebox{0.5}{0.3} &           37.6\scalebox{0.5}{0.0} &  \textbf{71.2\scalebox{0.5}{0.0}} &  \textbf{53.7\scalebox{0.5}{0.1}} &           30.8\scalebox{0.5}{0.1} &  3.02 \\
Influence (+)     &           75.9\scalebox{0.5}{0.0} &           67.7\scalebox{0.5}{0.2} &  \textbf{62.7\scalebox{0.5}{0.1}} &           53.2\scalebox{0.5}{0.2} &           52.9\scalebox{0.5}{0.3} &           38.1\scalebox{0.5}{0.1} &           70.6\scalebox{0.5}{0.1} &  \textbf{53.7\scalebox{0.5}{0.1}} &  \textbf{31.3\scalebox{0.5}{0.1}} &  \textbf{2.86} \\
\bottomrule

\newpage
~ & ~ & ~ & ~ & ~ & ~ & ~ & ~ & ~ & ~ & ~ \\
\textbf{OPT-13B} & ~ & ~ & ~ & ~ & ~ & ~ & ~ & ~ & ~ & ~ \\
\midrule
Similarity (+)    &           75.9\scalebox{0.5}{0.1} &           71.2\scalebox{0.5}{0.2} &           52.9\scalebox{0.5}{0.2} &           51.0\scalebox{0.5}{0.2} &           58.6\scalebox{0.5}{0.1} &           37.4\scalebox{0.5}{0.1} &           73.1\scalebox{0.5}{0.1} &           54.2\scalebox{0.5}{0.0} &           28.9\scalebox{0.5}{0.1} &  4.30 \\
Random            &           76.1\scalebox{0.5}{0.0} &           69.5\scalebox{0.5}{0.2} &           51.2\scalebox{0.5}{0.1} &           53.6\scalebox{0.5}{0.3} &           54.8\scalebox{0.5}{0.3} &           37.6\scalebox{0.5}{0.1} &           73.2\scalebox{0.5}{0.0} &           53.6\scalebox{0.5}{0.1} &           30.0\scalebox{0.5}{0.1} &  4.27 \\
One-shot (+)      &           75.8\scalebox{0.5}{0.0} &           69.0\scalebox{0.5}{0.1} &           57.3\scalebox{0.5}{0.2} &           50.0\scalebox{0.5}{0.0} &  \textbf{61.7\scalebox{0.5}{0.0}} &  \textbf{39.6\scalebox{0.5}{0.0}} &           72.4\scalebox{0.5}{0.0} &           53.3\scalebox{0.5}{0.1} &           32.0\scalebox{0.5}{0.0} &  4.17 \\
Perplexity (+)    &  \textbf{76.2\scalebox{0.5}{0.1}} &           71.8\scalebox{0.5}{0.1} &           55.9\scalebox{0.5}{0.2} &           53.1\scalebox{0.5}{0.3} &           42.4\scalebox{0.5}{0.2} &           38.0\scalebox{0.5}{0.0} &           73.1\scalebox{0.5}{0.0} &           54.1\scalebox{0.5}{0.1} &           28.0\scalebox{0.5}{0.1} &  4.14 \\
Best set          &           75.8\scalebox{0.5}{0.0} &  \textbf{72.8\scalebox{0.5}{0.1}} &           53.3\scalebox{0.5}{0.3} &           54.5\scalebox{0.5}{0.3} &           50.3\scalebox{0.5}{0.4} &           37.8\scalebox{0.5}{0.1} &           73.1\scalebox{0.5}{0.0} &           53.3\scalebox{0.5}{0.0} &           31.4\scalebox{0.5}{0.1} &  3.87 \\
IC Datamodels (+) &           75.9\scalebox{0.5}{0.0} &           72.0\scalebox{0.5}{0.1} &  \textbf{65.1\scalebox{0.5}{0.2}} &  \textbf{56.3\scalebox{0.5}{0.1}} &           48.1\scalebox{0.5}{0.4} &           37.2\scalebox{0.5}{0.0} &           72.6\scalebox{0.5}{0.0} &           54.3\scalebox{0.5}{0.0} &  \textbf{34.4\scalebox{0.5}{0.1}} &  3.38 \\
Influence (+)     &           75.8\scalebox{0.5}{0.0} &           71.9\scalebox{0.5}{0.1} &           61.7\scalebox{0.5}{0.2} &           55.7\scalebox{0.5}{0.1} &           57.1\scalebox{0.5}{0.2} &           36.8\scalebox{0.5}{0.1} &  \textbf{73.8\scalebox{0.5}{0.0}} &  \textbf{54.4\scalebox{0.5}{0.0}} &           34.0\scalebox{0.5}{0.1} &  \textbf{2.89} \\
\bottomrule

~ & ~ & ~ & ~ & ~ & ~ & ~ & ~ & ~ & ~ & ~ \\
\textbf{OPT-30B} & ~ & ~ & ~ & ~ & ~ & ~ & ~ & ~ & ~ & ~ \\
\midrule
Perplexity (+)    &           76.8\scalebox{0.5}{0.0} &           72.7\scalebox{0.5}{0.2} &           61.9\scalebox{0.5}{0.3} &           53.5\scalebox{0.5}{0.2} &           43.5\scalebox{0.5}{0.6} &           40.3\scalebox{0.5}{0.1} &           76.3\scalebox{0.5}{0.1} &           56.6\scalebox{0.5}{0.0} &           28.5\scalebox{0.5}{0.1} &  5.16 \\
Random            &           77.0\scalebox{0.5}{0.0} &           71.1\scalebox{0.5}{0.2} &           63.2\scalebox{0.5}{0.2} &           54.8\scalebox{0.5}{0.1} &           49.1\scalebox{0.5}{0.5} &           41.5\scalebox{0.5}{0.1} &           76.0\scalebox{0.5}{0.1} &           55.4\scalebox{0.5}{0.1} &           29.6\scalebox{0.5}{0.1} &  4.75 \\
Best set          &           76.9\scalebox{0.5}{0.0} &           72.6\scalebox{0.5}{0.0} &           64.1\scalebox{0.5}{0.3} &  \textbf{55.1\scalebox{0.5}{0.2}} &           54.8\scalebox{0.5}{0.4} &           40.8\scalebox{0.5}{0.0} &           75.8\scalebox{0.5}{0.1} &           56.1\scalebox{0.5}{0.0} &           31.5\scalebox{0.5}{0.0} &  4.30 \\
One-shot (+)      &           77.5\scalebox{0.5}{0.0} &           76.5\scalebox{0.5}{0.1} &           52.4\scalebox{0.5}{0.1} &           51.1\scalebox{0.5}{0.2} &  \textbf{61.6\scalebox{0.5}{0.0}} &           41.5\scalebox{0.5}{0.0} &           76.1\scalebox{0.5}{0.1} &           56.6\scalebox{0.5}{0.1} &           31.2\scalebox{0.5}{0.0} &  3.92 \\
Similarity (+)    &           77.7\scalebox{0.5}{0.1} &           70.1\scalebox{0.5}{0.4} &           63.9\scalebox{0.5}{0.1} &           53.3\scalebox{0.5}{0.1} &           57.1\scalebox{0.5}{0.7} &  \textbf{42.1\scalebox{0.5}{0.1}} &           76.2\scalebox{0.5}{0.1} &           56.7\scalebox{0.5}{0.0} &           29.3\scalebox{0.5}{0.0} &  3.84 \\
Influence (+)     &           78.0\scalebox{0.5}{0.0} &           74.1\scalebox{0.5}{0.1} &           64.6\scalebox{0.5}{0.1} &           52.5\scalebox{0.5}{0.1} &           51.4\scalebox{0.5}{0.3} &           41.6\scalebox{0.5}{0.1} &  \textbf{77.0\scalebox{0.5}{0.0}} &           57.4\scalebox{0.5}{0.0} &  \textbf{33.3\scalebox{0.5}{0.0}} &  2.89 \\
IC Datamodels (+) &  \textbf{78.1\scalebox{0.5}{0.0}} &  \textbf{77.0\scalebox{0.5}{0.0}} &  \textbf{65.9\scalebox{0.5}{0.1}} &           51.4\scalebox{0.5}{0.2} &           56.4\scalebox{0.5}{0.1} &  \textbf{42.1\scalebox{0.5}{0.0}} &           76.6\scalebox{0.5}{0.0} &  \textbf{58.2\scalebox{0.5}{0.0}} &           31.7\scalebox{0.5}{0.1} &  \textbf{2.41} \\
\bottomrule
\end{longtable}
\newpage
\begin{longtable}{llllllllllc}
\caption{Full results for negative selection across all models over 7 seeds.}
\label{tab_mainresults-neg-full} \\
\toprule
{} &                               PIQA &                            BoolQ &                                RTE &                               WIC &                              WSC & ARC-c & ARC-e & HS &                             OBQA &   Rank ($\downarrow$)\\
\midrule
\textbf{GPT-J-6B} & ~ & ~ & ~ & ~ & ~ & ~ & ~ & ~ & ~ & ~ \\
\midrule
Similarity (-)    &           76.0\scalebox{0.5}{0.0} &           63.2\scalebox{0.5}{0.2} &           53.5\scalebox{0.5}{0.2} &           55.3\scalebox{0.5}{0.2} &           48.9\scalebox{0.5}{0.5} &           38.2\scalebox{0.5}{0.0} &           73.5\scalebox{0.5}{0.0} &           49.6\scalebox{0.5}{0.0} &           27.2\scalebox{0.5}{0.0} &  5.51 \\
Random            &           75.7\scalebox{0.5}{0.1} &           61.7\scalebox{0.5}{0.3} &           56.2\scalebox{0.5}{0.3} &           51.9\scalebox{0.5}{0.3} &           48.5\scalebox{0.5}{0.4} &           37.7\scalebox{0.5}{0.1} &           73.3\scalebox{0.5}{0.0} &           49.1\scalebox{0.5}{0.1} &           27.6\scalebox{0.5}{0.1} &  4.98 \\
Worst set         &           76.2\scalebox{0.5}{0.0} &           60.2\scalebox{0.5}{0.1} &           52.6\scalebox{0.5}{0.1} &           52.1\scalebox{0.5}{0.2} &           48.7\scalebox{0.5}{0.3} &           37.7\scalebox{0.5}{0.0} &           71.8\scalebox{0.5}{0.1} &           48.5\scalebox{0.5}{0.1} &           27.1\scalebox{0.5}{0.0} &  4.22 \\
IC Datamodels (-) &           76.3\scalebox{0.5}{0.0} &           58.8\scalebox{0.5}{0.2} &           50.4\scalebox{0.5}{0.0} &           50.2\scalebox{0.5}{0.1} &           53.0\scalebox{0.5}{0.2} &           38.3\scalebox{0.5}{0.1} &           71.6\scalebox{0.5}{0.0} &           47.1\scalebox{0.5}{0.0} &  \textbf{24.9\scalebox{0.5}{0.0}} &  3.43 \\
Perplexity (-)    &  \textbf{73.1\scalebox{0.5}{0.1}} &           60.1\scalebox{0.5}{0.1} &           52.8\scalebox{0.5}{0.1} &           51.4\scalebox{0.5}{0.1} &           43.8\scalebox{0.5}{0.3} &  \textbf{37.1\scalebox{0.5}{0.1}} &           72.4\scalebox{0.5}{0.1} &  \textbf{46.3\scalebox{0.5}{0.0}} &           27.3\scalebox{0.5}{0.0} &  3.32 \\
Influence (-)     &           75.6\scalebox{0.5}{0.0} &  \textbf{57.4\scalebox{0.5}{0.2}} &           50.6\scalebox{0.5}{0.0} &  \textbf{48.4\scalebox{0.5}{0.1}} &           47.9\scalebox{0.5}{0.4} &           38.5\scalebox{0.5}{0.0} &  \textbf{71.1\scalebox{0.5}{0.0}} &           47.4\scalebox{0.5}{0.0} &           26.2\scalebox{0.5}{0.1} &  2.98 \\
One-shot (-)      &           76.4\scalebox{0.5}{0.0} &           58.7\scalebox{0.5}{0.2} &  \textbf{50.0\scalebox{0.5}{0.0}} &           50.0\scalebox{0.5}{0.0} &  \textbf{38.3\scalebox{0.5}{0.0}} &           37.4\scalebox{0.5}{0.1} &           72.8\scalebox{0.5}{0.1} &           46.5\scalebox{0.5}{0.1} &           25.3\scalebox{0.5}{0.1} &  \textbf{2.68} \\
\bottomrule

~ & ~ & ~ & ~ & ~ & ~ & ~ & ~ & ~ & ~ & ~ \\
\textbf{GPT-NeoX-20B} & ~ & ~ & ~ & ~ & ~ & ~ & ~ & ~ & ~ & ~ \\
\midrule
Similarity (-)    &           76.8\scalebox{0.5}{0.1} &           62.5\scalebox{0.5}{0.2} &           63.5\scalebox{0.5}{0.1} &           52.4\scalebox{0.5}{0.2} &           44.3\scalebox{0.5}{0.4} &           43.7\scalebox{0.5}{0.1} &           77.5\scalebox{0.5}{0.1} &           55.3\scalebox{0.5}{0.0} &           32.5\scalebox{0.5}{0.0} &  5.08 \\
Random            &           77.3\scalebox{0.5}{0.0} &           67.0\scalebox{0.5}{0.4} &           62.4\scalebox{0.5}{0.3} &           51.7\scalebox{0.5}{0.2} &           43.5\scalebox{0.5}{0.6} &           43.9\scalebox{0.5}{0.1} &           77.8\scalebox{0.5}{0.1} &           55.1\scalebox{0.5}{0.1} &           30.3\scalebox{0.5}{0.0} &  4.94 \\
Perplexity (-)    &  \textbf{76.3\scalebox{0.5}{0.0}} &           72.4\scalebox{0.5}{0.1} &           61.2\scalebox{0.5}{0.3} &           53.5\scalebox{0.5}{0.1} &           46.0\scalebox{0.5}{0.6} &           44.3\scalebox{0.5}{0.1} &           77.4\scalebox{0.5}{0.1} &  \textbf{52.6\scalebox{0.5}{0.0}} &           30.7\scalebox{0.5}{0.1} &  4.57 \\
Worst set         &           76.4\scalebox{0.5}{0.0} &           66.9\scalebox{0.5}{0.2} &           61.5\scalebox{0.5}{0.4} &           51.5\scalebox{0.5}{0.3} &           45.5\scalebox{0.5}{0.3} &           43.1\scalebox{0.5}{0.1} &           77.1\scalebox{0.5}{0.1} &           54.8\scalebox{0.5}{0.0} &           30.5\scalebox{0.5}{0.1} &  4.29 \\
One-shot (-)      &           77.4\scalebox{0.5}{0.0} &  \textbf{50.9\scalebox{0.5}{0.0}} &           62.2\scalebox{0.5}{0.2} &           50.0\scalebox{0.5}{0.0} &           60.9\scalebox{0.5}{0.0} &           42.4\scalebox{0.5}{0.1} &  \textbf{76.5\scalebox{0.5}{0.0}} &           52.8\scalebox{0.5}{0.0} &  \textbf{28.1\scalebox{0.5}{0.1}} &  3.13 \\
Influence (-)     &           76.7\scalebox{0.5}{0.1} &           53.9\scalebox{0.5}{0.1} &           58.1\scalebox{0.5}{0.3} &           50.8\scalebox{0.5}{0.1} &  \textbf{38.1\scalebox{0.5}{0.2}} &           41.8\scalebox{0.5}{0.1} &           76.6\scalebox{0.5}{0.0} &           53.9\scalebox{0.5}{0.0} &           29.2\scalebox{0.5}{0.0} &  2.68 \\
IC Datamodels (-) &           77.1\scalebox{0.5}{0.1} &           54.3\scalebox{0.5}{0.1} &  \textbf{57.7\scalebox{0.5}{0.3}} &  \textbf{49.4\scalebox{0.5}{0.1}} &           40.6\scalebox{0.5}{0.2} &  \textbf{41.4\scalebox{0.5}{0.1}} &           76.9\scalebox{0.5}{0.0} &           53.1\scalebox{0.5}{0.0} &           29.1\scalebox{0.5}{0.1} &  \textbf{2.52} \\
\bottomrule

~ & ~ & ~ & ~ & ~ & ~ & ~ & ~ & ~ & ~ & ~ \\
\textbf{LLaMA-7B} & ~ & ~ & ~ & ~ & ~ & ~ & ~ & ~ & ~ & ~ \\
\midrule
Random            &           78.0\scalebox{0.5}{0.1} &           80.3\scalebox{0.5}{0.1} &           63.4\scalebox{0.5}{0.2} &           51.1\scalebox{0.5}{0.2} &           41.7\scalebox{0.5}{0.3} &           44.3\scalebox{0.5}{0.0} &           78.7\scalebox{0.5}{0.1} &           58.8\scalebox{0.5}{0.1} &           32.3\scalebox{0.5}{0.0} &  5.16 \\
Similarity (-)    &           77.8\scalebox{0.5}{0.0} &           79.2\scalebox{0.5}{0.1} &           59.1\scalebox{0.5}{0.1} &           51.6\scalebox{0.5}{0.2} &           42.6\scalebox{0.5}{0.3} &           44.7\scalebox{0.5}{0.1} &           78.4\scalebox{0.5}{0.0} &           58.6\scalebox{0.5}{0.1} &           31.9\scalebox{0.5}{0.1} &  4.75 \\
Worst set         &           78.3\scalebox{0.5}{0.0} &           77.6\scalebox{0.5}{0.1} &           58.1\scalebox{0.5}{0.2} &           52.7\scalebox{0.5}{0.1} &           41.9\scalebox{0.5}{0.4} &           44.0\scalebox{0.5}{0.0} &           79.0\scalebox{0.5}{0.0} &           58.8\scalebox{0.5}{0.0} &           30.1\scalebox{0.5}{0.1} &  4.65 \\
One-shot (-)      &           78.4\scalebox{0.5}{0.0} &           78.7\scalebox{0.5}{0.1} &           69.1\scalebox{0.5}{0.1} &           51.5\scalebox{0.5}{0.1} &           61.8\scalebox{0.5}{0.0} &  \textbf{42.3\scalebox{0.5}{0.1}} &  \textbf{74.7\scalebox{0.5}{0.1}} &           57.7\scalebox{0.5}{0.0} &           32.0\scalebox{0.5}{0.1} &  4.40 \\
Perplexity (-)    &  \textbf{75.5\scalebox{0.5}{0.0}} &           81.3\scalebox{0.5}{0.1} &           57.9\scalebox{0.5}{0.2} &           50.7\scalebox{0.5}{0.1} &           41.1\scalebox{0.5}{0.5} &           43.0\scalebox{0.5}{0.1} &           77.7\scalebox{0.5}{0.0} &  \textbf{56.7\scalebox{0.5}{0.0}} &           28.7\scalebox{0.5}{0.1} &  3.13 \\
IC Datamodels (-) &           78.2\scalebox{0.5}{0.0} &           73.5\scalebox{0.5}{0.3} &           53.5\scalebox{0.5}{0.1} &           50.6\scalebox{0.5}{0.1} &           45.6\scalebox{0.5}{0.8} &           43.6\scalebox{0.5}{0.1} &           76.9\scalebox{0.5}{0.0} &           57.3\scalebox{0.5}{0.0} &           27.9\scalebox{0.5}{0.1} &  2.94 \\
Influence (-)     &           78.1\scalebox{0.5}{0.0} &  \textbf{71.9\scalebox{0.5}{0.1}} &  \textbf{52.7\scalebox{0.5}{0.1}} &  \textbf{49.3\scalebox{0.5}{0.1}} &  \textbf{40.4\scalebox{0.5}{0.4}} &           43.1\scalebox{0.5}{0.1} &           76.5\scalebox{0.5}{0.1} &           57.2\scalebox{0.5}{0.0} &  \textbf{27.2\scalebox{0.5}{0.1}} &  \textbf{2.16} \\
\bottomrule

~ & ~ & ~ & ~ & ~ & ~ & ~ & ~ & ~ & ~ & ~ \\
\textbf{LLaMA-13B} & ~ & ~ & ~ & ~ & ~ & ~ & ~ & ~ & ~ & ~ \\
\midrule
Similarity (-)    &           79.2\scalebox{0.5}{0.0} &           83.2\scalebox{0.5}{0.0} &           58.7\scalebox{0.5}{0.1} &           54.6\scalebox{0.5}{0.2} &           43.9\scalebox{0.5}{0.5} &           51.1\scalebox{0.5}{0.0} &           82.3\scalebox{0.5}{0.0} &           62.1\scalebox{0.5}{0.0} &           37.1\scalebox{0.5}{0.0} &  5.51 \\
Random            &           78.5\scalebox{0.5}{0.1} &           82.6\scalebox{0.5}{0.1} &           61.1\scalebox{0.5}{0.3} &           51.8\scalebox{0.5}{0.2} &           42.9\scalebox{0.5}{0.4} &           50.4\scalebox{0.5}{0.1} &           82.7\scalebox{0.5}{0.0} &           62.5\scalebox{0.5}{0.1} &           35.7\scalebox{0.5}{0.1} &  4.84 \\
Worst set         &           78.8\scalebox{0.5}{0.0} &           79.2\scalebox{0.5}{0.1} &           54.1\scalebox{0.5}{0.2} &           53.3\scalebox{0.5}{0.1} &           45.7\scalebox{0.5}{0.6} &           50.3\scalebox{0.5}{0.1} &           83.0\scalebox{0.5}{0.0} &           62.1\scalebox{0.5}{0.1} &           33.6\scalebox{0.5}{0.1} &  4.60 \\
Perplexity (-)    &  \textbf{74.9\scalebox{0.5}{0.0}} &           82.4\scalebox{0.5}{0.1} &           57.9\scalebox{0.5}{0.1} &           55.4\scalebox{0.5}{0.2} &           42.8\scalebox{0.5}{0.4} &           49.4\scalebox{0.5}{0.0} &  \textbf{81.4\scalebox{0.5}{0.0}} &  \textbf{58.7\scalebox{0.5}{0.0}} &           33.1\scalebox{0.5}{0.1} &  3.56 \\
One-shot (-)      &           78.7\scalebox{0.5}{0.0} &  \textbf{68.2\scalebox{0.5}{0.2}} &           53.9\scalebox{0.5}{0.1} &           53.1\scalebox{0.5}{0.1} &           55.4\scalebox{0.5}{0.7} &           50.0\scalebox{0.5}{0.1} &  \textbf{81.4\scalebox{0.5}{0.0}} &           61.0\scalebox{0.5}{0.0} &           26.1\scalebox{0.5}{0.1} &  3.22 \\
IC Datamodels (-) &           78.5\scalebox{0.5}{0.0} &           69.3\scalebox{0.5}{0.3} &  \textbf{50.0\scalebox{0.5}{0.0}} &           51.6\scalebox{0.5}{0.2} &  \textbf{38.9\scalebox{0.5}{0.1}} &           50.0\scalebox{0.5}{0.1} &           82.8\scalebox{0.5}{0.1} &           61.8\scalebox{0.5}{0.0} &  \textbf{22.0\scalebox{0.5}{0.1}} &  2.84 \\
Influence (-)     &           78.6\scalebox{0.5}{0.0} &           68.3\scalebox{0.5}{0.3} &  \textbf{50.0\scalebox{0.5}{0.0}} &  \textbf{50.6\scalebox{0.5}{0.2}} &           39.8\scalebox{0.5}{0.3} &  \textbf{49.3\scalebox{0.5}{0.1}} &           82.4\scalebox{0.5}{0.1} &           61.6\scalebox{0.5}{0.0} &           22.9\scalebox{0.5}{0.1} &  \textbf{2.43} \\
\bottomrule

~ & ~ & ~ & ~ & ~ & ~ & ~ & ~ & ~ & ~ & ~ \\
\textbf{OPT-6.7B} & ~ & ~ & ~ & ~ & ~ & ~ & ~ & ~ & ~ & ~ \\
\midrule
Similarity (-)    &           75.6\scalebox{0.5}{0.0} &           65.4\scalebox{0.5}{0.2} &           56.7\scalebox{0.5}{0.2} &           52.7\scalebox{0.5}{0.0} &           50.8\scalebox{0.5}{0.2} &           38.0\scalebox{0.5}{0.1} &           70.9\scalebox{0.5}{0.1} &           53.0\scalebox{0.5}{0.0} &           26.9\scalebox{0.5}{0.0} &  4.94 \\
Random            &           75.5\scalebox{0.5}{0.1} &           68.2\scalebox{0.5}{0.3} &           55.4\scalebox{0.5}{0.3} &           51.7\scalebox{0.5}{0.2} &           49.4\scalebox{0.5}{0.5} &           38.3\scalebox{0.5}{0.1} &           70.4\scalebox{0.5}{0.1} &           52.0\scalebox{0.5}{0.1} &           27.7\scalebox{0.5}{0.1} &  4.70 \\
Worst set         &           76.1\scalebox{0.5}{0.0} &           66.4\scalebox{0.5}{0.1} &           53.0\scalebox{0.5}{0.3} &           52.5\scalebox{0.5}{0.1} &           55.3\scalebox{0.5}{0.3} &           38.4\scalebox{0.5}{0.1} &           69.6\scalebox{0.5}{0.1} &           51.4\scalebox{0.5}{0.0} &           26.8\scalebox{0.5}{0.0} &  4.44 \\
Perplexity (-)    &  \textbf{75.1\scalebox{0.5}{0.0}} &           70.7\scalebox{0.5}{0.1} &           51.9\scalebox{0.5}{0.2} &           50.7\scalebox{0.5}{0.0} &           59.6\scalebox{0.5}{0.2} &           37.1\scalebox{0.5}{0.1} &  \textbf{69.5\scalebox{0.5}{0.0}} &  \textbf{47.8\scalebox{0.5}{0.0}} &           27.5\scalebox{0.5}{0.0} &  3.81 \\
Influence (-)     &           76.3\scalebox{0.5}{0.0} &  \textbf{61.9\scalebox{0.5}{0.3}} &           50.8\scalebox{0.5}{0.1} &           50.5\scalebox{0.5}{0.1} &           51.9\scalebox{0.5}{0.7} &           37.2\scalebox{0.5}{0.1} &           70.3\scalebox{0.5}{0.1} &           51.4\scalebox{0.5}{0.1} &           25.1\scalebox{0.5}{0.0} &  3.38 \\
One-shot (-)      &           76.0\scalebox{0.5}{0.0} &           65.1\scalebox{0.5}{0.2} &  \textbf{50.0\scalebox{0.5}{0.0}} &           50.0\scalebox{0.5}{0.0} &  \textbf{38.3\scalebox{0.5}{0.0}} &           37.6\scalebox{0.5}{0.1} &           71.2\scalebox{0.5}{0.0} &           50.6\scalebox{0.5}{0.1} &           26.8\scalebox{0.5}{0.1} &  3.13 \\
IC Datamodels (-) &           75.7\scalebox{0.5}{0.1} &           66.3\scalebox{0.5}{0.1} &           50.8\scalebox{0.5}{0.1} &  \textbf{48.1\scalebox{0.5}{0.2}} &           47.8\scalebox{0.5}{0.2} &  \textbf{36.9\scalebox{0.5}{0.1}} &           69.8\scalebox{0.5}{0.0} &           51.2\scalebox{0.5}{0.0} &  \textbf{23.6\scalebox{0.5}{0.1}} &  \textbf{2.65} \\
\bottomrule

\newpage
\textbf{OPT-13B} & ~ & ~ & ~ & ~ & ~ & ~ & ~ & ~ & ~ & ~ \\
\midrule
Random            &           76.1\scalebox{0.5}{0.0} &           69.5\scalebox{0.5}{0.2} &           51.2\scalebox{0.5}{0.1} &           53.6\scalebox{0.5}{0.3} &           54.8\scalebox{0.5}{0.3} &           37.6\scalebox{0.5}{0.1} &           73.2\scalebox{0.5}{0.0} &           53.6\scalebox{0.5}{0.1} &           30.0\scalebox{0.5}{0.1} &  5.10 \\
Similarity (-)    &           76.0\scalebox{0.5}{0.1} &           68.5\scalebox{0.5}{0.1} &           50.6\scalebox{0.5}{0.1} &           55.8\scalebox{0.5}{0.1} &           52.7\scalebox{0.5}{0.3} &           38.6\scalebox{0.5}{0.0} &           73.6\scalebox{0.5}{0.0} &           53.0\scalebox{0.5}{0.1} &           29.5\scalebox{0.5}{0.1} &  4.75 \\
Perplexity (-)    &  \textbf{73.7\scalebox{0.5}{0.1}} &           71.6\scalebox{0.5}{0.1} &           50.3\scalebox{0.5}{0.0} &           50.0\scalebox{0.5}{0.0} &           52.0\scalebox{0.5}{0.5} &           38.7\scalebox{0.5}{0.1} &  \textbf{72.5\scalebox{0.5}{0.1}} &           51.6\scalebox{0.5}{0.0} &           30.4\scalebox{0.5}{0.0} &  4.00 \\
IC Datamodels (-) &           77.0\scalebox{0.5}{0.0} &           69.1\scalebox{0.5}{0.2} &           50.3\scalebox{0.5}{0.0} &           50.9\scalebox{0.5}{0.1} &           50.7\scalebox{0.5}{0.3} &           36.7\scalebox{0.5}{0.1} &           72.6\scalebox{0.5}{0.1} &           53.1\scalebox{0.5}{0.0} &           28.8\scalebox{0.5}{0.1} &  3.84 \\
Worst set         &           76.1\scalebox{0.5}{0.0} &           67.2\scalebox{0.5}{0.2} &           50.4\scalebox{0.5}{0.0} &           50.7\scalebox{0.5}{0.1} &           48.8\scalebox{0.5}{0.4} &           37.4\scalebox{0.5}{0.1} &           72.9\scalebox{0.5}{0.0} &           53.1\scalebox{0.5}{0.1} &           28.6\scalebox{0.5}{0.0} &  3.83 \\
Influence (-)     &           76.6\scalebox{0.5}{0.0} &           69.4\scalebox{0.5}{0.1} &           50.4\scalebox{0.5}{0.1} &  \textbf{49.2\scalebox{0.5}{0.1}} &           45.4\scalebox{0.5}{0.3} &           36.5\scalebox{0.5}{0.1} &  \textbf{72.5\scalebox{0.5}{0.1}} &           52.7\scalebox{0.5}{0.0} &  \textbf{28.2\scalebox{0.5}{0.1}} &  3.05 \\
One-shot (-)      &           76.2\scalebox{0.5}{0.1} &  \textbf{52.1\scalebox{0.5}{0.0}} &  \textbf{50.0\scalebox{0.5}{0.0}} &           50.0\scalebox{0.5}{0.0} &  \textbf{38.3\scalebox{0.5}{0.0}} &  \textbf{35.7\scalebox{0.5}{0.1}} &  \textbf{72.5\scalebox{0.5}{0.1}} &  \textbf{50.2\scalebox{0.5}{0.0}} &           29.0\scalebox{0.5}{0.1} &  \textbf{2.14} \\
\bottomrule

~ & ~ & ~ & ~ & ~ & ~ & ~ & ~ & ~ & ~ & ~ \\
\textbf{OPT-30B} & ~ & ~ & ~ & ~ & ~ & ~ & ~ & ~ & ~ & ~ \\
\midrule
Similarity (-)    &           77.7\scalebox{0.5}{0.1} &           67.3\scalebox{0.5}{0.2} &           64.2\scalebox{0.5}{0.2} &           54.6\scalebox{0.5}{0.1} &           49.3\scalebox{0.5}{0.4} &           42.3\scalebox{0.5}{0.1} &           76.5\scalebox{0.5}{0.1} &           56.3\scalebox{0.5}{0.0} &           30.3\scalebox{0.5}{0.0} &  5.79 \\
Random            &           77.0\scalebox{0.5}{0.0} &           71.1\scalebox{0.5}{0.2} &           63.2\scalebox{0.5}{0.2} &           54.8\scalebox{0.5}{0.1} &           49.1\scalebox{0.5}{0.5} &           41.5\scalebox{0.5}{0.1} &           76.0\scalebox{0.5}{0.1} &           55.4\scalebox{0.5}{0.1} &           29.6\scalebox{0.5}{0.1} &  4.87 \\
Worst set         &           77.6\scalebox{0.5}{0.0} &           66.5\scalebox{0.5}{0.4} &           64.3\scalebox{0.5}{0.2} &           54.9\scalebox{0.5}{0.1} &           46.5\scalebox{0.5}{0.3} &           40.9\scalebox{0.5}{0.0} &           75.1\scalebox{0.5}{0.1} &           55.5\scalebox{0.5}{0.0} &           29.8\scalebox{0.5}{0.0} &  4.59 \\
Influence (-)     &           77.6\scalebox{0.5}{0.0} &           61.0\scalebox{0.5}{0.4} &           59.1\scalebox{0.5}{0.2} &           51.7\scalebox{0.5}{0.1} &           43.9\scalebox{0.5}{0.3} &           41.3\scalebox{0.5}{0.1} &           76.1\scalebox{0.5}{0.0} &           54.4\scalebox{0.5}{0.0} &           27.8\scalebox{0.5}{0.1} &  3.59 \\
Perplexity (-)    &  \textbf{75.1\scalebox{0.5}{0.0}} &           73.7\scalebox{0.5}{0.1} &           52.0\scalebox{0.5}{0.2} &           51.6\scalebox{0.5}{0.1} &           46.6\scalebox{0.5}{0.5} &           40.9\scalebox{0.5}{0.1} &  \textbf{74.5\scalebox{0.5}{0.1}} &           53.5\scalebox{0.5}{0.0} &           30.6\scalebox{0.5}{0.0} &  3.48 \\
IC Datamodels (-) &           77.4\scalebox{0.5}{0.1} &           59.1\scalebox{0.5}{0.1} &           60.1\scalebox{0.5}{0.3} &           51.1\scalebox{0.5}{0.1} &           42.8\scalebox{0.5}{0.2} &           40.4\scalebox{0.5}{0.1} &           75.4\scalebox{0.5}{0.1} &           55.2\scalebox{0.5}{0.0} &  \textbf{26.9\scalebox{0.5}{0.1}} &  2.98 \\
One-shot (-)      &           77.7\scalebox{0.5}{0.0} &  \textbf{51.7\scalebox{0.5}{0.0}} &  \textbf{50.0\scalebox{0.5}{0.0}} &  \textbf{50.0\scalebox{0.5}{0.0}} &  \textbf{38.3\scalebox{0.5}{0.0}} &  \textbf{40.3\scalebox{0.5}{0.0}} &           75.8\scalebox{0.5}{0.1} &  \textbf{51.7\scalebox{0.5}{0.0}} &           28.2\scalebox{0.5}{0.1} &  \textbf{2.02} \\
\bottomrule
\end{longtable}

\newpage

\begin{table}[t]
\caption{Prompt templates used in our experiments. Token `\texttt{\escape{n}\#\#\#\escape{n}}' is used to separate each example in few-shot prompting.}
\label{tab_prompts}
\vskip 0.1in
\centering
\begin{tabular}{ll}
\midrule
Task & Template\\
\midrule
PIQA & \multirow{2}{.7\textwidth}{Goal: \texttt{\{goal\}}\\ Answer: \texttt{\{answer\}}} \\
~ & ~ \\
~ & ~ \\
BoolQ & \multirow{2}{.7\textwidth}{\texttt{\{passage\}}\\ question: \texttt{\{question\}}?\\ answer: \texttt{\{answer\}}} \\
~ & ~ \\
~ & ~ \\
~ & ~ \\
RTE & \multirow{2}{.7\textwidth}{\texttt{\{premise\}}\\ question: \texttt{\{hypothesis\}}. true or false?\\ answer: \texttt{\{answer\}}} \\
~ & ~ \\
~ & ~ \\
~ & ~ \\
WIC & \multirow{3}{.7\textwidth}{\texttt{\{sentence1\}}\\ \texttt{\{sentence2\}}\\ question: Is the word `\texttt{\{word\}}' used in the same sense in the two sentences above?\\ answer: \texttt{\{answer\}}} \\
~ & ~ \\
~ & ~ \\
~ & ~ \\
~ & ~ \\
~ & ~ \\
WSC & \multirow{3}{.7\textwidth}{Passage: \texttt{\{text\}}\\ Question: In the passage above, does the pronoun `\texttt{\{span2\}}' refer to \texttt{\{span1\}}?\\ Answer: \texttt{\{answer\}}} \\
~ & ~ \\
~ & ~ \\
~ & ~ \\
~ & ~ \\
Arc (Chal.) & \multirow{2}{.7\textwidth}{Question: \texttt{\{question\}}\\ Answer: \texttt{\{answer\}}} \\
~ & ~ \\
~ & ~ \\
Arc (Easy) & \multirow{2}{.7\textwidth}{Question: \texttt{\{question\}}\\ Answer: \texttt{\{answer\}}} \\
~ & ~ \\
~ & ~ \\
Hellaswag & \multirow{2}{.7\textwidth}{Context: \texttt{\{context\}}\\ Answer: \texttt{\{answer\}}} \\
~ & ~ \\
~ & ~ \\
OBQA & \multirow{2}{.7\textwidth}{Context: \texttt{\{context\}}\\ Answer: \texttt{\{answer\}}} \\
~ & ~ \\
\bottomrule
\end{tabular}
\end{table}

\begin{table}[]
\caption{Highly positive influence examples from the top 20 percentile bin for LLaMA-7B.}
\label{tab_highprompts_llama}
\vskip 0.1in
\resizebox{\textwidth}{!}{
\begin{tabular}{llll}
\toprule
Task & ID & Influence & Prompt \\
\midrule
PIQA & 2305 & 0.004877 & \multirow{2}{.7\textwidth}{Goal: sand paper\\ Answer: can be used to smooth wood for furniture} \\
~ & ~ & ~ & ~ \\
~ & ~ & ~ & ~ \\
RTE & 1439 & 0.016923 & \multirow{4}{.7\textwidth}{As a real native Detroiter, I want to remind everyone that Madonna is from Bay City, Mich., a nice place in the thumb of the state's lower peninsula.\\ question: Madonna was born in Bay City, Mich.. true or false?\\ answer: true} \\
~ & ~ & ~ & ~ \\
~ & ~ & ~ & ~ \\
~ & ~ & ~ & ~ \\
~ & ~ & ~ & ~ \\
WIC & 2033 & 0.01273 & \multirow{5}{.7\textwidth}{Efface the memory of the time in the camps.\\ Efface oneself.\\ question: Is the word `efface' used in the same sense in the two sentences above?\\ answer: false} \\
~ & ~ & ~ & ~ \\
~ & ~ & ~ & ~ \\
~ & ~ & ~ & ~ \\
~ & ~ & ~ & ~ \\
~ & ~ & ~ & ~ \\
WSC & 98 & 0.031323 & \multirow{3}{.7\textwidth}{Passage: The man lifted the boy onto his bunk bed.\\ Question: In the passage above, does the pronoun `his' refer to The man?\\ Answer: false} \\
~ & ~ & ~ & ~ \\
~ & ~ & ~ & ~ \\
~ & ~ & ~ & ~ \\
Arc (Chal.) & 684 & 0.004829 & \multirow{2}{.7\textwidth}{Question: Which energy resource is non-renewable?\\ Answer: oil} \\
~ & ~ & ~ & ~ \\
~ & ~ & ~ & ~ \\
Arc (Easy) & 859 & 0.003275 & \multirow{2}{.7\textwidth}{Question: Which processes change magma into igneous rock?\\ Answer: cooling and crystallization} \\
~ & ~ & ~ & ~ \\
~ & ~ & ~ & ~ \\
Hellaswag & 30980 & 0.00546 & \multirow{8}{.7\textwidth}{Context: Education and Communications: [header] How to calculate consumer surplus [title] Understand the law of demand. [step] Most people have heard the phrase ``supply and demand'' used in reference to the mysterious forces governing market economies, but many don't understand these concepts' full implications. ``demand'' refers to the desire for a good or service in the marketplace.\\
Answer: Generally, if all other factors are equal, demand for a product will fall as its price increases. [substeps] For example, let's say that a company is about to release a new model of television.} \\
~ & ~ & ~ & ~ \\
~ & ~ & ~ & ~ \\
~ & ~ & ~ & ~ \\
~ & ~ & ~ & ~ \\
~ & ~ & ~ & ~ \\
~ & ~ & ~ & ~ \\
~ & ~ & ~ & ~ \\
~ & ~ & ~ & ~ \\
~ & ~ & ~ & ~ \\
OBQA & 3640 & 0.006248 & \multirow{2}{.7\textwidth}{Context: Scavengers eat dead what?\\ Answer: fauna} \\
 ~ & ~ & ~ & ~ \\
\bottomrule
\end{tabular}
}
\end{table}
\begin{table}[]
\caption{Highly negative influence examples from the bottom 20 percentile bin for LLaMA-7B.}
\label{tab_lowprompts_llama}
\vskip 0.1in
\resizebox{\textwidth}{!}{
\begin{tabular}{llll}
\toprule
Task & ID & Influence & Prompt \\
\midrule
PIQA & 10777 & -0.001315 & \multirow{2}{.7\textwidth}{Goal: baby wipe\\ Answer: Can be pierced by a fork Using the tines} \\
~ & ~ & ~ & ~ \\
~ & ~ & ~ & ~ \\
RTE & 2391 & -0.022086 & \multirow{4}{.7\textwidth}{Since the fear of death is virtually a universal phenomenon, the death penalty is an unparalleled deterrent for people considering a crime.\\ question: Capital punishment is a deterrent to crime.. true or false?\\ answer: true} \\
~ & ~ & ~ & ~ \\
~ & ~ & ~ & ~ \\
~ & ~ & ~ & ~ \\
~ & ~ & ~ & ~ \\
WIC & 4233 & -0.007281 & \multirow{5}{.7\textwidth}{After the fire a still small voice. -- 1 Kings 19:12.\\ Conservatism has many voices.\\ question: Is the word `voice' used in the same sense in the two sentences above?\\ answer: false} \\
~ & ~ & ~ & ~ \\
~ & ~ & ~ & ~ \\
~ & ~ & ~ & ~ \\
~ & ~ & ~ & ~ \\
~ & ~ & ~ & ~ \\
WSC & 334 & -0.007789 & \multirow{4}{.7\textwidth}{Passage: Sara borrowed the book from the library because she needs it for an article she is working on. She reads it when she gets home from work.\\ Question: In the passage above, does the pronoun `it' refer to the book?\\ Answer: true} \\
~ & ~ & ~ & ~ \\
~ & ~ & ~ & ~ \\
~ & ~ & ~ & ~ \\
~ & ~ & ~ & ~ \\
Arc (Chal.) & 596 & -0.006895 & \multirow{4}{.7\textwidth}{Question: A research scientist repeatedly observes a bird avoiding a specific butterfly species even  though it eats other types of butterflies. Which statement most likely explains the behavior of the bird?\\ Answer: The behavior is learned over the lifetime of the bird.} \\
~ & ~ & ~ & ~ \\
~ & ~ & ~ & ~ \\
~ & ~ & ~ & ~ \\
~ & ~ & ~ & ~ \\
Arc (Easy) & 1940 & -0.002372 & \multirow{3}{.7\textwidth}{Question: The organisms that convert solar energy and raw materials into food are\\ Answer: producers.} \\
 ~ & ~ & ~ & ~ \\
~ & ~ & ~ & ~ \\
~ & ~ & ~ & ~ \\
Hellaswag & 8891 & -0.005678 & \multirow{3}{.7\textwidth}{Context: Surfing: People are surfing on a large wave in the water. A boat is in the water. a large wave\\
Answer: crashes in the water.} \\
~ & ~ & ~ & ~ \\
~ & ~ & ~ & ~ \\
~ & ~ & ~ & ~ \\
OBQA & 978 & -0.004077 & \multirow{2}{.7\textwidth}{Context: Do objects change size with distance for Stevie Wonder?\\ Answer: No} \\
 ~ & ~ & ~ & ~ \\
\bottomrule
\end{tabular}
}
\end{table}
\end{document}